\documentclass[a4paper,fleqn]{cas-dc}
\usepackage[authoryear]{natbib}
\usepackage{subcaption}
\usepackage{multirow}
\usepackage{adjustbox}
\usepackage{amsmath,amsfonts,bm}
\usepackage{alphalph}
\usepackage{amssymb}
\usepackage{pifont}
\usepackage{comment}
\usepackage{graphicx}
\usepackage{booktabs}
\usepackage{array}
\usepackage{longtable}
\usepackage{microtype}
\usepackage{hyperref}
\usepackage{xcolor}
\usepackage{placeins}
\makeatletter
\renewcommand\paragraph{\@startsection{paragraph}{4}{\z@}%
  {3.25ex \@plus1ex \@minus.2ex}%
  {-1em}%
  {\normalfont\normalsize\bfseries}}
\let\cas@dbflt\@dbflt
\newcommand{\dblfloatbottom}{\def\@dbflt##1{\@xdblfloat{##1}[b]}}
\newcommand{\dblfloatdefault}{\let\@dbflt\cas@dbflt}
\makeatother

\newcommand{\model}{GeoPhysAdapter}

\begin{document}
\let\WriteBookmarks\relax
\renewcommand{\topfraction}{0.90}
\renewcommand{\bottomfraction}{0.80}
\renewcommand{\textfraction}{0.08}
\renewcommand{\floatpagefraction}{0.75}
\renewcommand{\dbltopfraction}{0.90}
\renewcommand{\dblfloatpagefraction}{0.75}
\emergencystretch=2em

\shorttitle{\model}
\shortauthors{Liu et~al.}
\title[mode = title]{GeoPhysAdapter: Scale-Matched Geophysical Adaptation for Cross-Domain Landslide Mapping with Vision Foundation Models}

\author[1]{Zhihang Liu}[type=editor, orcid=0009-0000-6281-418X]
\ead{zhihangliu@link.cuhk.edu.hk}

\author[1,2]{Mei-Po Kwan}[orcid=0000-0001-8602-9258]
\cormark[1]
\ead{mpk654@gmail.com}

\author[3]{Jinlin Wu}[orcid=0009-0009-3279-5374]
\ead{jwu923@connect.hkust-gz.edu.cn}

\author[4]{Hao Li}[orcid=0000-0002-6336-8772]
\ead{hao.li@nus.edu.sg}

\affiliation[1]{organization={Institute of Space and Earth Information Science, The Chinese University of Hong Kong},
 city={Shatin},
 postcode={999077},
 state={Hong Kong SAR},
 country={China}}

\affiliation[2]{organization={Department of Geography and Resource Management, The Chinese University of Hong Kong},
 city={Shatin},
 postcode={999077},
 state={Hong Kong SAR},
 country={China}}

\affiliation[3]{organization={Urban Governance and Design Thrust, The Hong Kong University of Science and Technology (Guangzhou)},
 city={Guangzhou},
 postcode={511400},
 state={Guangdong},
 country={China}}

\affiliation[4]{organization={Department of Geography, National University of Singapore},
 city={Singapore},
 postcode={117570},
 country={Singapore}}

\cortext[cor1]{Corresponding author}

\begin{abstract}
Newly triggered landslides rarely carry immediate annotations, so cross-domain transferability determines the value of landslide mapping for emergency response and regional risk assessment. Vision foundation models have strengthened representational transfer, yet on unseen regions, events, and data sources they still generate high-confidence false alarms. Terrain, material, and rainfall triggering can constrain such errors, but their supports are local, regional, and event-scale, so that resampling onto a 10~m grid misaligns them with the segmentation decision unit and compounds the uncertain geographic context problem (UGCoP). We propose GeoPhysAdapter, which anchors on a frozen vision foundation model, restricts terrain, material, and triggering to dense spatial guidance, regional modulation, and event-timing forcing, and applies bounded adaptation at two decision units, the pixel and the candidate landslide body, reverting exactly to the visual prediction where support is insufficient. On an event-isolated PILD dataset of four public sources, 55 global landslide events, and 7,890 test samples, 70.3\% of cross-domain false-positive mass lies in near-pure spurious bodies of median equivalent diameter 207~m, matching coarse-prior support rather than the pixel. Pixel-level adaptation removes a net 507,817 erroneous pixels and reduces error by 7.76\%, whereas raising the decision unit to the candidate body, under identical samples, anchor, and baseline, increases error reduction to 23.99\%, approximately 3.1 times the pixel-level effect, improves IoU by 0.031 (14.2\% relative), and corrects 9.92 pixels per pixel harmed. The gain decreases monotonically with terrain displacement, exceeds a strong optical control of full spectral change and visual confidence alone by 0.019 IoU, and remains positive across five vision anchors. A pre-registered single-shot re-execution preserves 0.027 IoU and a 20.61\% error reduction. What constrains this correction is therefore the match between the native scale of physical evidence and the operative decision unit, rather than denser physical stacking alone. The data and code are publicly available at \url{https://github.com/Liu-Zhihang/geophysadapter}.
\end{abstract}


\maketitle

\section{Introduction}

\begin{figure*}[!t]
\centering
\includegraphics[width=\textwidth]{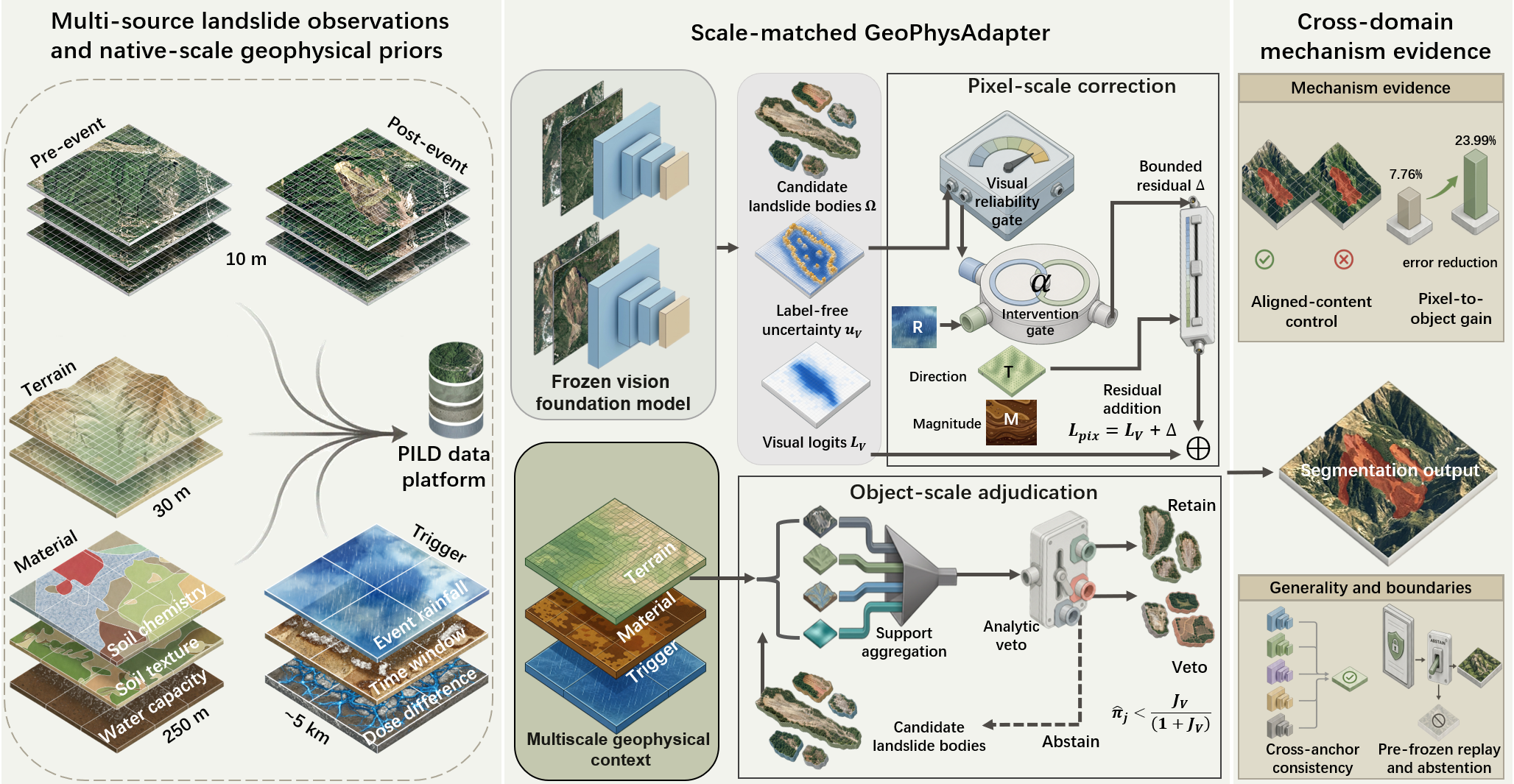}
\caption{Conceptual overview of GeoPhysAdapter. Left: multi-source optical observations and the three geophysical priors under a unified PILD input contract. Centre: a frozen vision foundation model emits $L_V$, $u_V$, and candidate bodies $\Omega$; pixel-scale adaptation applies a bounded residual $L_{\mathrm{pix}}=L_V+\Delta$, while object-scale review retains or vetoes support-valid candidates under $\hat{\pi}_j<J_V/(1+J_V)$ and abstains otherwise. Right: matched evidence that the gain depends on aligned physical content and that raising the decision unit from pixel to candidate body widens error reduction from 7.76\% to 23.99\%.}
\label{fig:overview}
\end{figure*}

Accurate and fast landslide mapping from remote sensing is essential for post-disaster emergency response, landslide inventory updating, and regional risk assessment \citep{Casagli2023}. Benchmark standardization, deep segmentation, and multi-temporal observation have substantially improved pixel-level landslide delineation \citep{Ghorbanzadeh2022,Bhuyan2023}, and large-scale pretraining with adapter-style transfer has brought vision foundation models into transferable landslide mapping \citep{Oquab2023,Wei2024,Lu2025}. In-benchmark performance, however, is not equivalent to cross-domain reliability. When a model is transferred to a new region, event, or data source, geomorphic setting, surface spectra, imaging conditions, and annotation conventions may change together, so that the visual cues formed during training no longer hold. The consequence is not only a lower overall score, but also missed landslides, high-confidence false alarms over non-landslide terrain, and miscalibrated predicted probabilities \citep{Hong2023,Rafi2024,Minderer2021}. Newly occurring events rarely carry immediate annotations, which makes it impractical to retrain or recalibrate a model for every scene. Reliable cross-domain transfer is therefore a precondition for trustworthy GeoAI in disaster-response settings \citep{Wang2024,Janowicz2020,Hochmair2025}.

This cross-domain fragility points to a more fundamental misalignment in landslide mapping. Remote-sensing imagery records the surface appearance left behind after failure, whereas landsliding is a physical process controlled jointly by terrain, the properties of surface material, and hydrometeorological triggering \citep{Guzzetti2008,Sidle2016,Reichenbach2018,Liu2023}. Within this triad, terrain describes where failure is spatially predisposed, material describes the medium in which failure can develop, and triggering describes the temporal forcing that brings an event about; together they encode spatial variation in terrain, surface material, and rainfall forcing. In principle, these three geophysical priors can constrain cross-domain visual errors, but their native supports differ by orders of magnitude: among globally available products, terrain is typically resolved at about 30~m, soil properties at about 250~m, and event rainfall at about 5~km \citep{ESA2022,Poggio2021,Funk2015}. Resampling these layers onto a common 10~m analysis grid changes only how the arrays are spatially aligned; it does not create physical information at that scale. When coarse context is broadcast directly to a pixel-level segmentation head, its genuine contribution may be diluted, and the model may instead exploit source-correlated shortcuts \citep{Geirhos2020,Rafi2024}. The uncertain geographic context problem (UGCoP; \citealp{Kwan2012a,Kwan2012b}) states this risk in general terms: the proxy geographic context available to a model need not coincide, in spatial support, temporal window, or alignment fidelity, with the setting that actually governed failure. For cross-domain landslide mapping, UGCoP is therefore not merely data noise but the gap between ``a physical channel is available'' and ``that evidence can legitimately correct the present visual errors.'' Physical relevance alone cannot guarantee effective pixel-level adaptation; trustworthy physical intervention presupposes support fidelity, role constraints, and matching of the decision scale. This leads to the question that motivates the present study: can coarse geophysical priors, once scale-matched, adaptively correct the cross-domain landslide-mapping errors of vision foundation models in a trustworthy and interpretable way?

Research addressing this question has shifted from strengthening visual representations toward incorporating domain knowledge, yet the decisive chain remains open. Vision foundation models and domain adaptation improve representational transfer, but their residual errors remain predominantly appearance-driven \citep{Oquab2023,Cong2022,Wei2024,Lu2025}. Theory-guided and physics-informed learning, together with landslide susceptibility research, establish the explanatory value of terrain, material, and triggering, yet typically take the slope, region, or event as the analysis unit rather than the 10~m post-event boundary \citep{Karpatne2017,Reichstein2019,Karniadakis2021,Reichenbach2018,Dahal2025}. Object-based remote-sensing analysis shows that the decision unit constrains which contextual evidence can act \citep{Blaschke2010}, but has not resolved how priors of differing support should enter a vision foundation model under cross-domain transfer. More critically, foundation-model transfer, physically informed learning, and cross-domain reliability are usually evaluated under separate protocols: studies that compare adaptation scales under one visual anchor and one test corpus, and that contrast aligned with misaligned physical content, remain scarce \citep{Reichstein2019,Lu2025}. The central open question is therefore no longer whether more physical channels can be stacked onto a model, but how physical information of differing support can intervene at a decision unit commensurate with its reach, while preserving probabilistic reliability and scientific interpretability.

Motivated by this gap, we propose GeoPhysAdapter, a scale-matched geophysical adaptation framework for vision foundation models. The core design principle is not to replace a pre-trained visual encoder with a separate physical segmenter, but to treat its prediction as an anchor and to apply bounded correction according to the role, support scale, and support quality of each prior. The vision foundation model first produces visual logits, a label-free uncertainty, and candidate landslide bodies. Terrain, as the only prior that can be densely aligned in space, supplies geometric direction; material and triggering, which represent a static susceptibility background and an event-scale temporal forcing, modulate adaptation where their support is valid and abstain where it is not. GeoPhysAdapter then operates at two complementary scales: at the pixel scale, predictive reliability limits a bounded residual update that targets scattered visual errors; at the object scale, visual confidence and validity-tested geophysical support are aggregated within each connected component so that a candidate body is retained, vetoed, or abstained on as a whole. The object decision follows an analytically derived improvement criterion rather than a threshold searched on test data, and the adapter reverts exactly to the visual prediction wherever physical information is missing, misaligned, or unreliable, thereby turning UGCoP into an executable intervene-or-abstain contract. Figure~\ref{fig:overview} provides a conceptual overview of the framework and verification claims; the operators and module interfaces are given in Figure~\ref{fig:framework}.

To test whether the framework genuinely uses physical content, rather than additional capacity, data-source identity, or the choice of threshold, we build a verification scheme matched to these methodological claims (Figure~\ref{fig:overview}). The experiments use a unified, event-isolated PILD corpus from four public sources and compare pixel-level with object-level adaptation under a fixed visual anchor, fixed test samples, and a fixed baseline threshold. Aligned, spatially shifted, spatially rolled, and cross-event donor controls test whether the gain depends on physically correct content. Pixel-level evidence spans eight vision architectures; the object-level mechanism is replayed on five vision anchors; and a pre-frozen execution tests whether the development observations survive without rule adjustment. Terrain, material, and triggering are additionally examined on their native tasks and scales, so that the existence, usability, and transfer of physical information into landslide segmentation can be distinguished.

This study makes four main contributions. First, starting from the tension between the object structure of cross-domain visual errors and the coarse support of geophysical priors, it operationalizes UGCoP as an adaptation principle of support fidelity, role constraint, and scale matching. Second, it proposes the unified GeoPhysAdapter framework, in which role separation, bounded updating, object-level aggregation, and active abstention allow geophysical priors to intervene without replacing the vision foundation model. Third, it establishes a pixel-versus-object comparison under identical samples, an identical visual anchor, and an identical baseline on an event-isolated multi-source dataset, and attributes the gain through misalignment controls, cross-architecture replication, and a pre-frozen execution. Fourth, it examines separately the native-task information of terrain, material, and triggering and the boundary of its transfer to landslide segmentation, thereby specifying when physical adaptation should be applied and when the visual judgment should be retained. Taken together, GeoPhysAdapter answers how geophysical priors can be used, why physical information is able to act, and where the limits of that action lie, providing a unified framework for interpretable cross-domain landslide mapping within trustworthy GeoAI.

\section{Related Work}

The literature most relevant to this study falls into four connected lines of work: cross-domain landslide mapping, physically informed learning, vision foundation models, and object-based remote-sensing analysis. Each has substantially advanced visual representation transfer, the interpretation of landslide processes, or the modeling of spatial structure \citep{Casagli2023,Lu2025,Ma2017OBIA}, but each is normally pursued under a different task and a different evaluation protocol. One question of direct relevance to the present study therefore remains unanswered: when the visual model, the test corpus, and the baseline prediction are all held fixed, at which decision unit should geophysical priors of differing support intervene in order to correct cross-domain visual errors, and how can the resulting gain be shown to come from aligned physical content?

\subsection{Appearance-Based Transfer and Structured Failure in Cross-Domain Landslide Mapping}

Benchmark standardization, deep segmentation, and multi-temporal remote sensing have moved landslide mapping from single-event manual interpretation toward repeatable dense prediction \citep{Casagli2023,Ghorbanzadeh2022,Bhuyan2023}. As the setting has shifted from in-benchmark testing toward unseen regions and events, methodological attention has followed it into cross-domain representation learning. Prototype-guided domain-aware representation learning reduces the feature discrepancy between source and target domains \citep{Zhang2023CrossDomain}, freezing a pre-trained segmentation model and training a lightweight adapter lowers the parameter and annotation cost of regional transfer \citep{Wei2024}, and transfer learning from vision foundation models supplies reusable landslide representations \citep{Hou2025VFM}. Taken together, these studies show that feature alignment, parameter-efficient adaptation, and pre-trained representations can all mitigate cross-region appearance change.

Existing transfer strategies, however, draw mainly on target-image statistics, a small number of target labels, or correspondences between visual features, so that the basis for correction still lies in image appearance \citep{Zhang2023CrossDomain,Wei2024,Hou2025VFM}. Where geomorphic setting, imaging conditions, and annotation conventions change together, a model may produce spatially coherent false alarms over visually similar bare ground, river bars, roads, or mining areas, rather than independent random pixel errors. General cross-domain remote-sensing segmentation and trustworthy remote-sensing studies converge on the same conclusion, in that domain shift degrades segmentation accuracy and predictive reliability at the same time \citep{Hong2023,Rafi2024,Wang2024}. Cross-view disaster mapping further shows that heterogeneous observation geometry can leave event damage incompletely recovered from a single view \citep{Li2025CrossView}, and complementary work that fuses volunteered geographic information with remote-sensing evidence likewise shows that multi-source geographic context can expose omissions that imagery alone does not resolve \citep{Li2020OSM}. Visual transfer has thus answered how representations can be reused, but not what non-visual information can constrain the errors that remain when visual evidence fails systematically. This situation has motivated researchers to look beyond appearance-level transfer and toward geophysical domain knowledge for structured remedies.

\subsection{Susceptibility Explanation and Boundary Correction in Landslide Science}

From a geoscience perspective, landslide science supplies an explanatory system that does not rest on visual appearance. Statistical susceptibility studies have long used terrain, lithology, soil, and hydrometeorological variables to characterize the conditions of failure, and have established the stable explanatory value of these factors for landslide occurrence probability \citep{Reichenbach2018}. Theory-guided data science and physics-informed machine learning argue further that explicit incorporation of domain knowledge can improve extrapolation, scientific consistency, and interpretability \citep{Karpatne2017,Reichstein2019,Karniadakis2021}. Within landslide research, physically informed learning has begun to connect susceptibility assessment, slope-stability reasoning, and data-driven prediction \citep{Liu2023,Dahal2025}.

Methodologically, however, this body of work explains why a given region, slope, or event is more likely to fail, whereas post-event landslide mapping has to determine the specific boundary that the present event has left behind, and the two do not share a decision unit \citep{Reichenbach2018,Dahal2025}. Terrain can impose a continuous geometric constraint locally, soil and lithology usually describe a coarser and largely static medium, and rainfall mainly supplies temporal forcing at the scale of the event. Even where these variables are genuinely related to the failure process, resampling them and broadcasting them onto 10~m pixels does not by itself produce boundary evidence at that scale. The UGCoP makes the same point from the side of context, in that the geographic proxies available to a model may depart from the spatial and temporal setting that actually acts on the object of study \citep{Kwan2012a,Kwan2012b}. Existing physical research therefore establishes which factors are related to failure, without resolving how those factors should act in post-event dense prediction \citep{Reichstein2019,Karniadakis2021,Casagli2023}. We accordingly move the problem from unconditional physical fusion toward the matching of role and scale: a prior should enter the visual prediction only where its support is commensurate with the decision unit of the error to be corrected.

\subsection{Transferable Representations and Reliable Adaptation in Vision and Earth Observation Foundation Models}

Large-scale self-supervised pretraining allows foundation models to learn transferable visual representations in the absence of task labels. DINOv2 provides general-purpose visual features \citep{Oquab2023}, while SatMAE, SpectralGPT, and SkySense++ build remote-sensing pretraining capacity from multi-temporal, multispectral, and multimodal Earth observation respectively \citep{Cong2022,Hong2024,Wu2025}. Prithvi-EO-2.0 further supplies a multi-temporal Earth-observation foundation model with native multispectral inputs and temporal--location embeddings, and has demonstrated transferable representations across diverse remote-sensing tasks \citep{Szwarcman2024}; we adopt its 300M-TL variant as the primary visual anchor in Sections~4--5. Surveys of remote-sensing and geo-foundation models likewise document rapid expansion toward multimodal and domain-general settings, while also stressing remaining gaps between representational capacity and trustworthy geographic deployment \citep{Lu2025,Janowicz2025}. On this basis, parameter-efficient adaptation has entered transferable landslide mapping, and vision foundation models are beginning to be used for landslide segmentation and change detection \citep{Wei2024,Hou2025VFM,Leonardi2025}. Compared with single-task networks trained from scratch, such models provide a stronger and reusable visual anchor for comparison across data sources.

Representational capacity is not equivalent to reliability under geographic extrapolation. Recent work on Earth foundation models identifies scale awareness, the expression of uncertainty, and physical consistency as remaining conditions for reliable Earth-system inference \citep{Zhu2026EarthFM,Wu2025}. Broader GeoAI discussions emphasize spatially explicit reasoning and responsible practice \citep{Janowicz2020,Hochmair2025}, together with reproducible evaluation under geographic shift \citep{Li2024a}. The calibration and selective-prediction literature adds that model confidence is not naturally equivalent to error risk, so that a reliable system has to restrain its predictions on high-risk inputs or be allowed to reject them \citep{Guo2017,Minderer2021}; recent remote-sensing work likewise couples progressive uncertainty guidance with binary segmentation \citep{Li2026}. Existing adaptation methods mostly adjust features or a small number of parameters. They seldom distinguish the roles that dense spatial evidence, coarse background, and event-level forcing play in a decision, and they usually lack a direct contrast between physically aligned and physically misaligned input \citep{Wei2024,Hou2025VFM,Lu2025}. A stronger visual anchor can therefore raise the baseline without answering, on its own, when a physical prior should intervene, how far it should intervene, and whether it should withdraw when the evidence is insufficient. We treat the foundation model as a visual judgment to be preserved, rather than as a fusion carrier that coarse priors may overwrite unconditionally.

\subsection{Decision Units and Scale Matching from Pixels to Geographic Objects}

Object-based remote-sensing analysis grew out of a reconsideration of the per-pixel paradigm, in that pixels within one geographic entity are not independent of one another, and their spectral, shape, textural, neighborhood, and semantic relations acquire their full meaning only at an appropriate spatial unit \citep{Blaschke2010,Blaschke2014}. GEOBIA accordingly organizes image segmentation, spatial relations, and multi-source geographic attributes at the object level, and treats scale as part of the definition of an object rather than as a fixed property of the input \citep{Blaschke2014,Ma2017OBIA}. Studies of multiresolution segmentation show further that object scale directly affects internal heterogeneity and boundary integrity, and must therefore be selected from scene structure or local variance rather than set arbitrarily \citep{Dragut2010,Ma2017OBIA}. This literature establishes a principle of immediate relevance here, namely that the unit of analysis determines which spatial and contextual information a model can use effectively.

Object-based reasoning has a clear line of development within landslide mapping. Early work combined spectral, shape, adjacency, and terrain-morphological evidence to identify candidate landslide objects and to exclude spectrally similar features such as river sand and bare rock \citep{Martha2010}. Segmentation-scale optimization and data-driven thresholding subsequently improved the stability with which the boundaries of differently sized landslide objects could be delineated \citep{Martha2011}, and random forests with object-level feature selection reduced the dependence of hand-crafted rules on region and sensor \citep{Stumpf2011}. Multi-temporal object analysis then brought brightness change, texture, context, and digital terrain models together to compile historical landslide inventories \citep{Martha2012}, while hierarchical multiresolution approaches exploited the nesting relations among spatial resolutions and used terrain information to guide the segmentation and classification of landslide objects \citep{Kurtz2014}. Collectively, this work demonstrates that landslides should not be treated only as mutually independent pixels, and that candidate objects can carry morphological, contextual, and terrain evidence which is difficult to express stably at the pixel level.

Existing object-level landslide methods, however, usually generate and classify objects directly from high-resolution imagery of a particular region, and the scale to which they refer is mainly a segmentation parameter or a resolution level \citep{Martha2010,Martha2011,Stumpf2011,Kurtz2014}. The problem addressed here is different, in that a vision foundation model has already produced a cross-domain prediction, and what has to be decided is whether coarse geophysical support should correct that prediction at the pixel or at the candidate landslide body. Object analysis, foundation model adaptation, and cross-domain landslide mapping still lack systematic evidence comparing the two adaptation scales on one test corpus and one visual prediction, and they rarely use spatial shifts, spatial rolls, or cross-event donors to test whether a gain depends on physically correct content \citep{Ma2017OBIA,Zhang2023CrossDomain,Wei2024,Hou2025VFM}. Without controls of this kind, the smoothing effect of object aggregation, the effect of additional model capacity, and a genuine geophysical contribution remain difficult to separate.

Taken together, this literature delimits where GeoPhysAdapter enters. It is not a further model of unconditional multimodal fusion, but a way of organizing geophysical intervention around a fixed visual anchor: terrain, material, and triggering enter the pixel decision or the candidate-object decision according to their own role and support scale, the magnitude of any update is constrained by predictive reliability and by the quality of physical support, and the adapter abstains where the evidence is insufficient. Accordingly, we do not take whether additional physical channels raise average accuracy as the only test. We instead pose three mutually constraining and falsifiable questions: whether the gain depends on aligned physical content, whether the decision scale determines the strength of correction that transfers, and whether the adapter can avoid negative transfer where effective support is absent.

\section{Data and Problem Setting}

This study defines cross-domain landslide mapping as binary segmentation under multi-source, event-isolated conditions. The data are organized on two principles. The visual input, the label, and the prediction grid must obey a single contract, whereas every geophysical input retains its true provenance, its native support scale, and its quality state, and is not reinterpreted as an observation at the analysis scale merely because it has been resampled. The first principle keeps model comparisons consistent, and the second provides the auditable basis for scale matching and for physical attribution.

\subsection{The Unified Four-Source PILD Corpus}

The main experiments are built on the geo-ready four-source cohort of PILD (Physics-Informed Landslide Dataset), which comprises DLR Landslide Reference \citep{Martinis2025}, GDCLD \citep{Fang2024}, GLaD4CD v1 \citep{Leonardi2024}, and Sen12Landslides harmonized \citep{Hoehn2025}. Every sample carries an auditable spatial reference, an image window, a landslide mask, and an event identity. After label-independent quality control on image integrity, temporal availability, and geolocation, the frozen corpus contains 7,890 samples and 55 canonical events; the source-level composition, QC exclusions, and event-merge rule are given in Supplementary Table~S1.

\begin{figure*}[!t]
\centering
\includegraphics[width=\textwidth]{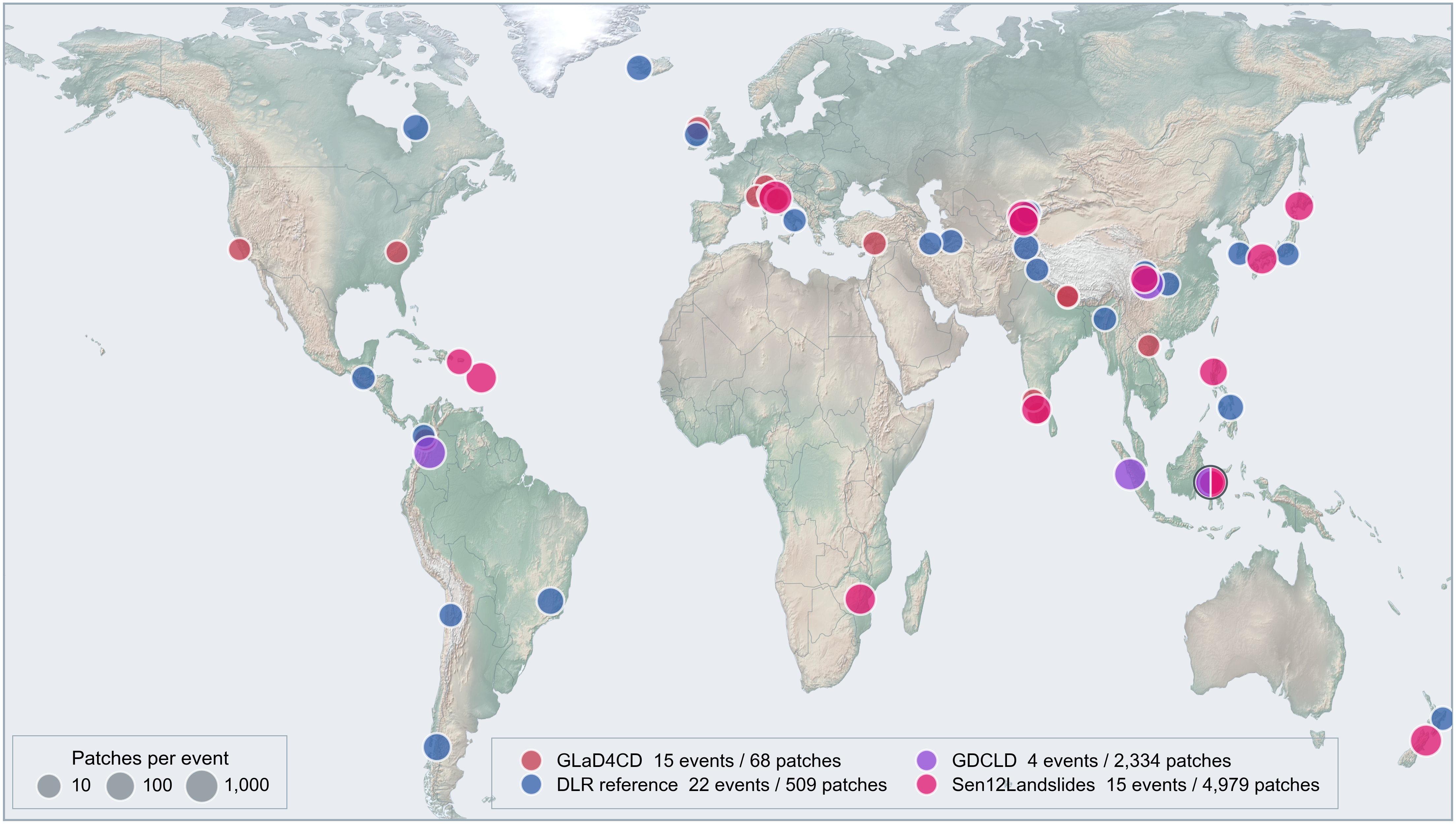}
\caption{Spatial distribution of the 55 canonical events in the unified four-source PILD corpus. Marker color denotes data source and marker area scales logarithmically with samples per event; the two-color Palu marker is the single cross-source merge used before splitting.}
\label{fig:map}
\end{figure*}

The within-source event counts sum to 56 while the canonical count is 55, because one set of spatiotemporally coincident records in GDCLD and Sen12Landslides was identified as a single canonical event and is always assigned to the same data role. Sample density differs markedly across sources: GLaD4CD v1 contributes only about 4.5 samples per event on average, so it serves as a cross-region stress test and its event-wise results are interpreted at the event level rather than the sample level. The CAS Landslide data \citep{Xu2024} remain registered in PILD, but their samples lack an auditable coordinate reference system (CRS) and affine transform, so no pixel-level terrain correspondence can be established and they do not enter the spatial physical-attribution corpus used here. One further DLR event of 26 samples was excluded because its four-date availability and valid optical coverage fell below a pre-frozen data-quality threshold. None of these decisions read any segmentation result. Figure~\ref{fig:map} shows the global distribution of the canonical events across the four sources.

\subsection{Optical Inputs, Labels, and the Analysis Grid}

All main experiments use a common six-band Sentinel-2 L2A input, namely B02, B03, B04, B8A, B11, and B12. Each sample contains the two nearest pre-event dates and the two earliest post-event dates, four dates in total, and retains acquisition dates, the latitude and longitude of the sample center, the valid-optics mask, and cloud metadata. Imagery is represented on a $128\times128$ grid aligned to 10~m, corresponding to a fixed spatial footprint of $1.28\times1.28$~km. The 10~m here refers only to the grid on which the model performs its analysis and on which the labels are aligned; it does not change the native information scale of any geophysical product.

The data contract thereby establishes a scale discipline that runs through the whole study: each class of input is first derived and summarized at its own native support, and only then aligned to the shared 10~m analysis grid. Whether a class of information can take part in a per-pixel decision depends on whether spatial variation usable for localization survives within the sample after resampling, not on whether it has been resampled at all. This discipline yields a hierarchy that can be counted directly. Within a 1.28~km sample, the 10~m optical grid has 128 cells per side, the 20~m bands 64, the 30~m terrain about 43, the 250~m material about 5, and event rainfall at about 5~km fewer than one. The first three retain sufficient within-sample variation and can therefore participate in per-pixel discrimination, whereas the last two cannot and consequently take on background and event-level roles within the framework.

Landslide annotations from the different sources are harmonized into binary masks, and a valid-pixel mask is retained alongside them; invalid regions enter neither the loss nor the metrics. Sample identity is fixed by \texttt{sample\_id}, and event identity by \texttt{canonical\_event\_id} after cross-source deduplication. Differences between sources in observation availability and in original annotation conventions do not disappear entirely under resampling, so the source is retained only as an audit variable and an optional conditioning variable, and is never treated as physical evidence. Standardization parameters for all continuous optical and geophysical variables are estimated on the current training fold alone.

\subsection{The Three Geophysical Priors and Their Native Scales}

GeoPhysAdapter uses terrain, material, and triggering priors, but does not assume that they share the same spatial semantics. Terrain (CopDEM GLO-30 and 17 morphometric derivatives; native support about 30~m with 90--900~m windows) is the sole dense spatial direction. Material (21 soil/moisture attributes at about 250~m) enters only as a bounded amplitude multiplier in $[0.75,1.25]$ and never as an independent boundary. Triggering (three CHIRPS pre-event rainfall contrasts at about 5~km) supplies an event-level intervention dose and never an independent direction. The role overview and the complete variable inventory, derivation windows, and invalid-support behavior are given in Supplementary Tables~S2a--S2b and in the text below.

All terrain derivations are computed on the native, approximately 30~m equidistant grid in the target projection within a 3~km buffer, and only then reprojected onto the 10~m analysis grid. This order of derivation is frozen with the cache as a property of the dataset, so the 90--900~m window statistics correspond to genuine geomorphic support rather than to a false resolution obtained by resampling repeatedly on a fine grid. Local, mesoscale, and macroscale variables enter three separate terrain branches, which preserves their support scales. The material variables come from OpenLandMap available water capacity and SoilGrids 250~m soil properties \citep{Poggio2021}, and the triggering variables from CHIRPS daily rainfall \citep{Funk2015}. One 250~m material cell covers 625 pixels of 10~m and therefore cannot indicate which of them failed, and event rainfall at about 5~km has no spatial variation within a single sample, so that its per-pixel information content is identically zero. Both facts are geometric and do not change with the modeling choice, which is why material and triggering can enter the framework only as background and event-level modulation.

Each prior is accompanied by an independent support-quality variable, $q_T$, $q_M$, and $q_R$. Here $q_T$ records per-pixel DEM validity and the coverage quality of the sample, while $q_M$ and $q_R$ record whether the material and event-timing evidence meet the provenance, completeness, and timing thresholds. In the object-analysis corpus, valid material and triggering support covers 77.8\% and 45.6\% of the candidate bodies respectively. Missing or invalid support is never substituted by a pooled mean posing as an observation; the corresponding quality variable is instead set exactly to zero, which returns the model to the state in which that prior is not used.

Source-level quality control in PILD, the complete variable inventory, and the behavior under invalid support are summarized in Supplementary Tables~S1 and~S2a--S2b.

\subsection{Event-Isolated Splits, Freezing, and Auxiliary Cohorts}

The main corpus uses a four-fold, source-stratified, event-isolated split (the split contract and hash freeze are recorded in Supplementary Table~S1). Within each fold, any canonical event serves exactly one of training, validation, or testing, and the cross-source observations and all samples of one event are always bound together as a single unit. The four outer test folds are mutually exclusive and their union covers all events, so that after merging, each of the 7,890 samples receives exactly one test prediction. Training uses temperature-controlled source-event-sample stratified sampling to reduce the dominance of large events and large sources, whereas testing preserves the natural sample proportions and reports pooled, event-macro, and per-source results side by side.

Content hashes are recorded before training for the unified manifest, the event alias table, the split files, the optical cache, the terrain cache, and the Material and Trigger registries. Any missing asset, inconsistent identity ordering, or changed hash triggers a refusal to execute. Model selection and threshold selection use only the corresponding outer training and validation data, and outer test labels take no part in input standardization, candidate generation, routing fits, or threshold determination.

Beyond the main PILD corpus, three auxiliary cohorts answer different attribution questions. The 800 official test samples of Landslide4Sense support a deterministic pixel-level replay across eight vision architectures, which tests whether the effect of terrain depends on a single backbone \citep{Ghorbanzadeh2022}. GLaD events, after spatial exclusion against the PILD development events, are used separately for the external native-task tests of terrain susceptibility and of material interaction. The temporal attribution of triggering uses 138 independent events with reliable event dates and rainfall coverage, merged into 90 storm clusters by a 7-day time window and a 250~km spatial proximity relation. Landslide4Sense provides no usable event identifier under this split, so its evidence is used only for cross-architecture consistency and not for any claim of cross-event generalization. These auxiliary cohorts are never scored jointly with the main PILD results; they serve only to verify the information content of the three priors, their consistency across architectures, and the boundary of their applicability.

\section{GeoPhysAdapter}

\begin{figure*}[!t]
\centering
\includegraphics[width=\textwidth]{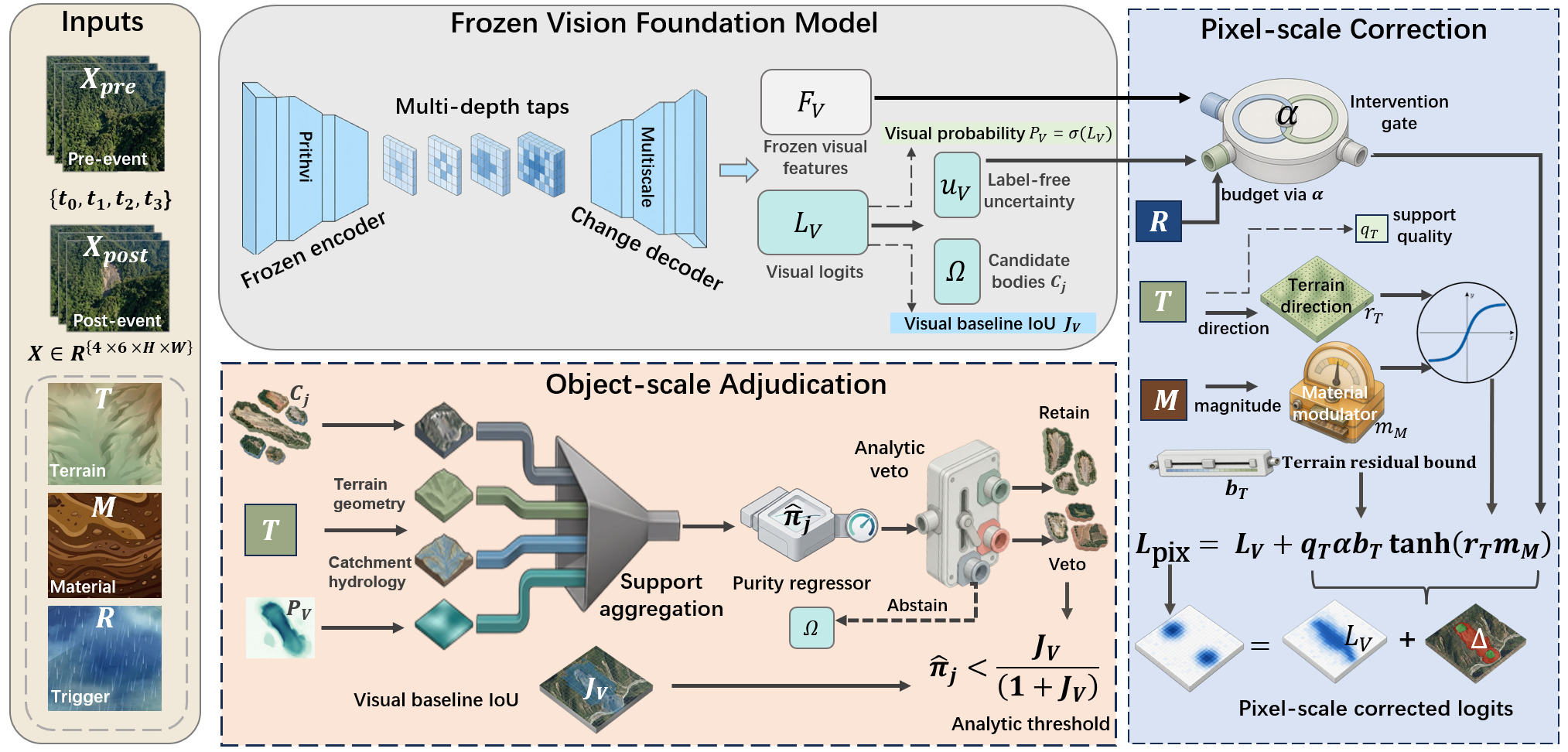}
\caption{Scale-matched technical framework of GeoPhysAdapter. Left: optical input $X$ with priors $T$, $M$, and $R$. Upper centre: frozen Prithvi encoder and change decoder yield $F_V$, $L_V$, $u_V$, $P_V=\sigma(L_V)$, and candidates $\Omega/C_j$. Right: pixel update $L_{\mathrm{pix}}=L_V+q_T\alpha b_T\tanh(r_T m_M)$ with identity fallback where support is invalid. Lower centre: 92-D object descriptors feed purity $\hat{\pi}_j$; support-valid candidates are vetoed when $\hat{\pi}_j<J_V/(1+J_V)$ and otherwise retained.}
\label{fig:framework}
\end{figure*}

GeoPhysAdapter is designed as a constrained intervention around a vision foundation model, rather than as a physical segmentation pathway able to replace the visual prediction on its own. Its unifying principle is that the visual model is responsible for proposing visual logits and candidate landslide bodies from the remote-sensing imagery, whereas the physical modules may correct that visual judgment only where support is valid and where the decision scale is compatible with the physical information, and otherwise revert exactly to the original visual output. The principle is realized at two parallel, rather than sequential, decision scales. Figure~\ref{fig:overview} has already given the conceptual overview; this section and Figure~\ref{fig:framework} present the operator-level framework. The pixel level tests whether dense terrain support can apply a bounded correction to the visual logits, while material and triggering only modulate residual magnitude and the intervention budget respectively; the object level aggregates terrain geometry, catchment hydrology, spectral change, and visual confidence within each visual candidate, retains or vetoes support-valid candidates under an analytic threshold, and abstains where support is invalid.

\subsection{Task Formulation, Inputs, and the Visual Anchor}

Let $X\in\mathbb{R}^{4\times6\times H\times W}$ denote the four-date, six-band optical input, $Y\in\{0,1\}^{H\times W}$ the landslide label, and $L_V=f_\theta(X)$ the logit map produced by the visual anchor. The three physical inputs are written as a terrain tensor $T$, a material background vector $M$, and an event triggering vector $R$, with support qualities $q_T(x)$, $q_M$, and $q_R$. The pixel-level output of GeoPhysAdapter is

\begin{equation}
L_{\mathrm{pix}}(x)=L_V(x)+\Delta(x),
\end{equation}

where

\begin{equation}
\begin{aligned}
\Delta(x)
&=q_T(x)\,
  \alpha\!\left(F_V(x),u_V(x);R,q_R\right)\\
&\quad{}\times b_T\tanh\!\left[r_T(T)(x)\,
  m_M(M,q_M)\right],\\[0.3em]
m_M
&=1+\rho_M\,q_M\tanh\!\left(h_M(M)\right),
\qquad \rho_M=0.25.
\end{aligned}
\end{equation}

Here $r_T$ reads the terrain input alone and returns a spatial direction of correction. The gate $\alpha$ uses the frozen visual features $F_V$, the label-free visual uncertainty $u_V=1-|2\sigma(L_V)-1|$, and the event-level triggering scalar that calibrates the intervention budget to determine where intervention is permitted. The scalar $b_T$ bounds the maximum change in logit.

This form writes the scale differences among the three priors directly into the structure of the operator, so that each prior enters only where its native scale permits. Terrain is the only quantity with per-pixel localizing ability and is therefore the sole directional term inside the $\tanh$. Material can act only as a bounded positive multiplier on the magnitude of that direction, with a range of $[0.75,1.25]$, and can neither reverse the direction nor create one where the direction is zero. Triggering is a scalar that is constant within an event and has no within-sample variation available for localization, so it calibrates only the intervention budget expressed by $\alpha$, that is, how large a physical intervention this event should be allowed, while the location itself remains determined entirely by the frozen visual uncertainty together with the terrain direction.

The operator has exact fallback behavior. When $q_M=0$, $m_M\equiv1$; when $q_R=0$, $\alpha$ falls back to a purely visual gating form; and when $q_T(x)=0$, $\Delta(x)\equiv0$ and the pixel reverts identically to the visual prediction. Because $\Delta$ carries $q_T$ and the terrain direction as factors throughout, no value of triggering or material can produce a correction where terrain gives no direction. Missing support is therefore never turned by a fill value into something treated as observed physics, and abstention is a built-in behavior of the operator rather than a post hoc deletion of variables.

The main visual anchor is Prithvi-EO-2.0-300M-TL \citep{Szwarcman2024}. The model natively accepts the six bands B02, B03, B04, B8A, B11, and B12, and encodes Earth-observation sequences using temporal and positional coordinates. We extract multi-depth representations from Transformer blocks 6, 12, 18, and 24, construct pre-event, post-event, and absolute-change features explicitly at each depth, and then fuse them through a compact multi-scale decoder and a full-resolution spectral-change stem to obtain $L_V$ and $F_V$. The main experiments freeze the pre-trained encoder and train only the visual change decoder, and all physical experiments load the visual checkpoint of the same fold, which keeps the visual representation, the optical input, and the threshold unchanged.

Freezing the encoder keeps foundation-model gains separate from geophysical adaptation and prevents misaligned-terrain controls from rewriting the visual representation through joint training. Dependence on Prithvi is tested by repeating the object-level protocol on four alternative anchors (Section~5).

\subsection{Role-Constrained Pixel-Level Adaptation}

The terrain encoder enforces a strict separation in which support produces direction and vision decides timing. The 17 terrain variables enter zero-preserving convolutions and ConvNeXt-style residual blocks in fine, meso, and macro groups, which extract local hillslope form, mesoscale slope position, and macroscale valley-ridge context at full, $1/2$, and $1/4$ resolution respectively. The three branches are fused from coarse to fine and finally produce the single-channel terrain direction $r_T(T)$. Convolution biases and affine normalization are removed from the network, and the zero-terrain response is explicitly subtracted, so that a zero input maps strictly to zero; visual features cannot enter the terrain direction branch.

The material module produces no independent spatial map. From the 21-dimensional medium background it predicts low-rank, bounded Feature-wise Linear Modulation (FiLM)-style coefficients \citep{Perez2018} and applies them to scale-grouped bases of the terrain response (slope, curvature, relief), so that material can only rescale an existing terrain correction.

The triggering module aggregates the three event rainfall quantities into an event-level scalar that calibrates the intervention budget of the visual uncertainty gate. All modules are initialized from the identity state and are fail-closed under $q_M$ and $q_R$.

The visual decoder is trained with a positively weighted binary cross-entropy together with a soft Dice loss. The physical stage freezes the visual model and, beyond the same segmentation objective, applies a preservation term on the pixels that the original visual model already classified correctly, together with an $L_1$ constraint on the total change in logit:

\begin{equation}
\begin{aligned}
\mathcal{L}_{\mathrm{adapt}}
&=\mathcal{L}_{\mathrm{BCE}}+0.5\mathcal{L}_{\mathrm{Dice}}\\
&\quad{}+0.1\mathcal{L}_{\mathrm{preserve}}
+10^{-3}\lVert L_{\mathrm{pix}}-L_V\rVert_1.
\end{aligned}
\end{equation}

To prevent the router from merely memorizing in-distribution training errors, the PILD pixel experiments generate nested event-isolated out-of-fold (OOF) predictions within each outer fold. The help and harm labels of a candidate physical action are generated only from inner test predictions on the outer training events, and the utility gate learns from these whether to accept a correction; the gate parameters and the execution receipt are frozen before the outer test data are constructed. Any action that is not accepted restores $L_V$ bitwise, and outer test labels never take part in candidate generation or in fitting the gate.

The frozen parameters of visual training, pixel adaptation, the utility gate, and the object purity regression are reported in Supplementary Table~S3.

\subsection{Candidate Landslide Bodies and Object-Level Description}

Pixel-level adaptation still requires a coarse physical prior to make a local decision on a 10~m grid. To test the influence of the decision scale directly, we decompose the binary prediction obtained from the visual anchor at the threshold chosen on the validation set into eight-connected components $C_j$, and define each connected component as one candidate landslide body. The object-level adapter may only retain or veto a candidate body as a whole; vetoed pixels are restored to the visual negative class, and the adapter creates no new landslide in regions where the visual model predicted none.

This asymmetry follows from physical logic and from measurement alike. Failure to satisfy the hillslope and hydrological setting can constitute evidence for a veto, whereas satisfying the predisposing conditions is not by itself proof that a landslide occurred in the present event, because gravitational constraint is necessary rather than sufficient. The opposite direction was tested to the same standard: promoting an adjacent candidate body that the visual model had missed has a label-informed ceiling of $\Delta\mathrm{IoU}$ $+0.056$, but a deployable ranking correlation of only $0.17$--$0.22$, which is insufficient for deployment and was therefore formally excluded. Physical review is thus unidirectional by design, and this is a conclusion rather than a simplification.

The abstention rule at the object level is symmetric with the one at the pixel level. If the proportion of valid terrain support within a candidate body falls below a pre-frozen threshold, or if its hydrological and spectral descriptors cannot be computed completely within the valid mask, the candidate body does not enter the veto decision and is retained as it is, which is an identical reversion to the visual prediction.

Each candidate body is represented by a 92-dimensional descriptor comprising four complementary groups: (i)~terrain-geometry (27 dimensions), including area, compactness, elongation, aspect consistency, downslope direction, slope quantiles, terrain relief, topographic position index (TPI), valley-bottom and ridge-line position, curvature, and roughness; (ii)~visual-confidence (3 dimensions), namely the mean, maximum, and 90th-percentile landslide probability within the candidate body; (iii)~spectral-change (39 dimensions), comprising pre-event, post-event, difference, within-body heterogeneity, and outer-ring contrast of the Sentinel-2 bands that the segmentation network has already observed, together with the normalized difference vegetation index (NDVI), normalized burn ratio (NBR), normalized difference water index (NDWI), and bare-soil index (BSI); and (iv)~catchment-hydrology (23 dimensions), comprising DEM-derived D8 (eight-direction) flow accumulation, distance to channel, height above channel, slope position, and their candidate-body-to-outer-ring contrasts.

Both the terrain and the hydrological descriptors are aggregated over the candidate body and its spatial neighborhood, which converts support native to 30--900~m into statistical evidence commensurate with the candidate body. The spectral and confidence groups are retained as a strong visual competitor, used to separate the effectiveness of objectification itself from the additional information supplied by correct geophysical content. The main configuration further adds the data-source identifier known at deployment time in order to absorb residual annotation differences, while a source-blind configuration that does not use the identifier is evaluated alongside it.

\subsection{Event-Isolated Purity Estimation and Analytic Veto}

The true purity of a candidate body is defined as

\begin{equation}
\pi_j=\frac{|C_j\cap Y|}{|C_j|}.
\end{equation}

We predict $\hat{\pi}_j$ from the object descriptors with a HistGradientBoosting regressor. Training and prediction use a five-fold GroupKFold on \texttt{canonical\_event\_id}, so that all candidate bodies of one event always lie in the same fold, and predictions from five fixed random seeds are averaged to reduce the stochasticity of the tree model. Every candidate body therefore receives a purity score from a model that has never seen the event to which it belongs. Labels are used only to construct the purity regression target on the training folds and for the final offline evaluation, and take no part in computing the descriptors of test candidate bodies.

The object-level decision threshold is determined directly from an algebraic condition on IoU, rather than searched on the test results. Let $\mathrm{TP}$, $\mathrm{FP}$, and $\mathrm{FN}$ denote the numbers of true-positive, false-positive, and false-negative pixels of the visual prediction in the current evaluation pool, whose baseline IoU is

\begin{equation}
J_V=\frac{\mathrm{TP}}{\mathrm{TP}+\mathrm{FP}+\mathrm{FN}}.
\end{equation}

If a candidate body $C_j$ contains $i_j$ true-positive and $f_j$ false-positive pixels, then removing it as a whole gives

\begin{equation}
J'_j=\frac{\mathrm{TP}-i_j}{\mathrm{TP}+\mathrm{FP}+\mathrm{FN}-f_j}.
\end{equation}

From $J'_j>J_V$ it follows that $i_j/f_j<J_V$. Using $\pi_j=i_j/(i_j+f_j)$, the condition for a beneficial veto of a candidate body is equivalent to

\begin{equation}
\pi_j<\frac{J_V}{1+J_V}.
\end{equation}

GeoPhysAdapter therefore vetoes a candidate body when $\hat{\pi}_j<J_V/(1+J_V)$ and retains it otherwise. The condition is a marginal criterion relative to a fixed visual baseline, applied independently to each candidate body without sequential re-estimation, and all reported metrics are measured on the full pool after every veto has been executed, so that the conclusions do not depend on the marginal approximation. Because $J_V$ varies with the performance of the visual anchor itself, the criterion adapts to the error level of that anchor: the stronger the anchor, the higher the threshold and the wider the purity range within which a veto remains beneficial. Development results use the pooled baseline $J_V$ of the corresponding anchor, a quantity that is itself development information, whereas the pre-registered single-shot execution derives $J_V$ strictly from the fitting partition alone and no statistic of the held-out partition enters the decision. In neither case does the object-level method contain an additional tunable threshold. The final output is

\begin{equation}
\widehat{Y}_{\mathrm{obj}}=
\bigcup_{j:\,\hat{\pi}_j\ge J_V/(1+J_V)} C_j.
\end{equation}

Pixel and object branches are evaluated under the same visual prediction and pixel metric, so their contrast isolates the decision unit.

\section{Experimental Setup}

The setup asks whether geophysical support can correct visual judgments at the pixel and object scales, whether each prior carries native-task information, and whether reported gains are attributable to physical content rather than backbone or development choices. Unless otherwise noted, every adaptation result is compared pairwise with the visual prediction on the same fold, the same visual state, and the same test samples.

\subsection{Main Training and Cross-Prediction Protocol}

So that pixel-level and object-level adaptation share one visual starting point, the main PILD analysis first fixes a visual anchor and then evaluates both adaptation scales on its predictions. The main anchor is the official Prithvi-EO-2.0-300M-TL checkpoint \citep{Szwarcman2024}, chosen because it matches the common Sentinel-2 six-band, four-date contract and supplies temporal--location embeddings, not to rank foundation models. The encoder is frozen and the change decoder is trained for 30 epochs (AdamW, batch size 16, learning rate $3\times10^{-4}$, weight decay $10^{-4}$, gradient clipping at 1; source--event stratified sampling with temperature 0.75). Full optimizer and learner settings are collected in Supplementary Table~S3.

PILD uses four-fold canonical-event-isolated cross-prediction: each event is an outer test event exactly once, and its imagery, labels, and candidate-body statistics enter neither training, threshold selection, nor adapter fitting. The visual threshold is fixed on the outer validation fold and then applied to both the visual baseline and the matched adapter. A single pre-fixed Prithvi seed locks the visual prediction so that pixel--object differences can be attributed to adaptation scale rather than retraining noise. Statistical independence is always the canonical event; same-anchor seed repeats are robustness checks only. Cross-model robustness uses five visual anchors, and the Sen12 confirmation uses five seeds with five whole-region holdout folds.

Object-level purity regression uses event-grouped five-fold cross-prediction and a five-seed HistGradientBoosting ensemble on the 92-dimensional candidate descriptor; the source-blind control removes only the source identifier. Descriptor set, learner, analytic criterion, and their configuration were settled in development, and each candidate is scored only by a regressor that has never seen its event (Supplementary Table~S3).

These choices enforce three nested isolation levels against test-information leakage. Outer event isolation keeps each test event out of fitting and thresholding. Nested cross-prediction inside the outer training partition forms help/harm labels for the utility gate so that acceptance is not fit on in-distribution training errors; unaccepted actions restore $L_V$ bitwise, and gate parameters are frozen before outer-test construction. Pre-frozen single-shot execution, reported in Section~6.5, seals descriptors, learner, seeds, criterion, partition rule, and input hashes before a held-out re-run. The third level constrains analytical freedom rather than claiming prospective independent validation, because held-out labels were accessed in earlier data development.

\subsection{Visual Baselines and the Cross-Architecture Matrix}

Published results for public methods differ from this protocol in splits, channels, budget, threshold, and metrics, so Landslide4Sense reference systems, TransLandSeg, and Universal Adapter serve only as state-of-the-art background, not a numerical leaderboard. Attributable comparisons come only from matched visual--adapter pairs inside this study.

The pixel-level matrix covers DINOv2-S, ConvNeXtV2-FCMAE, Hiera-S-MAE, SatMAE-ViT-B, DeepLabV3+, U-Net++, FPN, and ImageNet-pretrained DeepLabV3+ \citep{Chen2018,He2016}, each with five seeds on all 800 official Landslide4Sense test samples, sharing a tensor-identical visual state and the visual system's threshold. The object-level cross-anchor set uses Prithvi-EO-2.0-300M-TL, DINOv3-SAT-L, Hiera-S-MAE, ConvNeXtV2-FCMAE, and DINOv2-S under a shared PILD manifest, split, sampling, budget, and evaluation code, but each anchor keeps its own threshold, baseline IoU, and analytic veto bound; we test directional replication rather than an average gain across unequal baselines.

\subsection{Attribution Controls and Ablations}

A gain after adding a channel is not physical attribution. We therefore install complementary controls:
\begin{enumerate}
\item Terrain content: standardized zero terrain, a 320~m zero-filled shift, a 640~m spatial roll, and a same-source cross-event donor, with candidates, spectra, and visual confidence held fixed;
\item Material: aligned Material, same-source cross-event shuffle, and $q_M=0$ identity fallback;
\item Triggering: aligned pre-event rainfall, same-location time shifts, cross-event shuffle, and $q_R=0$ identity fallback;
\item Strong optical competitors: confidence alone, spectral change alone, spectra plus confidence, and the same learner with terrain-geometry or catchment-hydrology descriptors added;
\item Source: source-conditioned versus source-blind object models, plus whole-source holdout for transfer to a new source;
\item Scale: pixel versus object adaptation on the same 7,890 samples, Prithvi prediction, baseline, and pixel metric.
\end{enumerate}

Native-task probes for terrain, material, and triggering are role-matched; when native information fails misalignment controls in segmentation, the prior retains native evidence and the segmentation adapter abstains (Supplementary Tables~S4--S6).

\subsection{Evaluation Metrics, Statistical Units, and Evidence Tiers}

The primary segmentation metric is foreground Intersection over Union (IoU). Average precision (AP) measures threshold-free ranking, and the Brier score and negative log-likelihood (NLL) measure probability quality. Visual error reduction is
\begin{equation}
\mathrm{RER}=
\frac{(\mathrm{FP}_V+\mathrm{FN}_V)-(\mathrm{FP}_A+\mathrm{FN}_A)}
{\mathrm{FP}_V+\mathrm{FN}_V}.
\end{equation}
Here $V$ and $A$ denote the visual and adapted outputs. A \texttt{corrected} pixel is wrong under the visual model and right after adaptation; a \texttt{harmed} pixel is the reverse. In an object veto, cleared false-positive pixels count as corrected and true-positive pixels deleted with the candidate as harmed. Matched hard decisions share the visual threshold; a more favorable adapter-only test threshold is not allowed.

Main results report pooled metrics at natural proportions, event-macro metrics, and per-source breakdowns together: the pool measures error removed under the actual composition, whereas the event macro prevents large events from substituting sample count for independent event evidence. Independence in PILD is the canonical event, with event or event-cluster bootstrap intervals. Landslide4Sense has no reliable event ID, so five-seed predictions are aggregated within each of the 800 samples before paired inference (never treated as 4,000 independent samples) and multi-architecture tests use Holm correction. Sen12 uses a seed--region hierarchical bootstrap; Trigger inference uses the 90 storm clusters.

Evidence is graded as \texttt{DEVELOPMENT} (event-grouped OOF on opened folds), \texttt{PRE-REGISTERED SINGLE-SHOT} (frozen re-execution; Section~6.5), or \texttt{BOUNDARY} (prespecified control failure or unstable cross-event/source effect). The single-shot split is not prospective validation, because held-out labels were seen in earlier data development.

\dblfloatbottom
\begin{figure*}
\centering
\includegraphics[width=\textwidth]{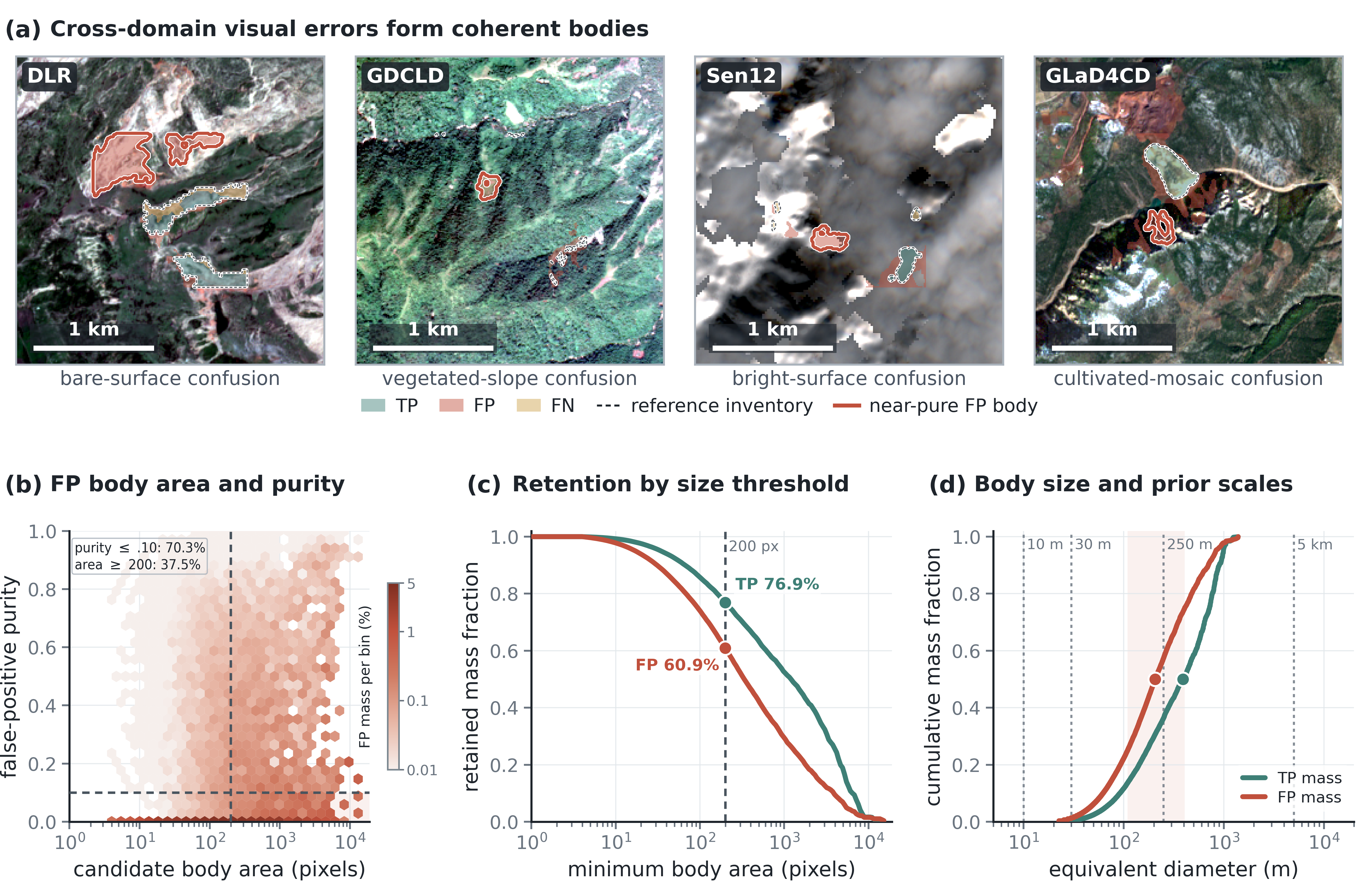}
\caption{Object structure of cross-domain visual errors. Teal, coral, and pale gold denote true-positive, false-positive, and missed pixels. (a)~Four out-of-fold samples from DLR, GDCLD, Sen12Landslides, and GLaD4CD with near-pure spurious bodies (purity $\le0.10$, area $\ge200$ pixels) outlined in coral; scale bar 1~km. (b)~False-positive mass in the area--purity plane: purity $\le0.10$ carries 70.3\%. (c)~Retained TP/FP mass versus minimum candidate area (76.9\%/60.9\% at 200 pixels). (d)~Mass-weighted equivalent-diameter CDF with native support scales of the 10~m pixel, 30~m Terrain, 250~m Material, and $\sim$5~km Trigger; shaded band is the false-positive IQR (109--407~m). Panel-selection criteria for panel~(a) follow Section~6.1 (purity $\le0.10$, area $\ge200$ pixels).}
\label{fig:errorstruct}
\end{figure*}
\dblfloatdefault

\dblfloatbottom
\begin{figure*}
\centering
\includegraphics[width=\textwidth]{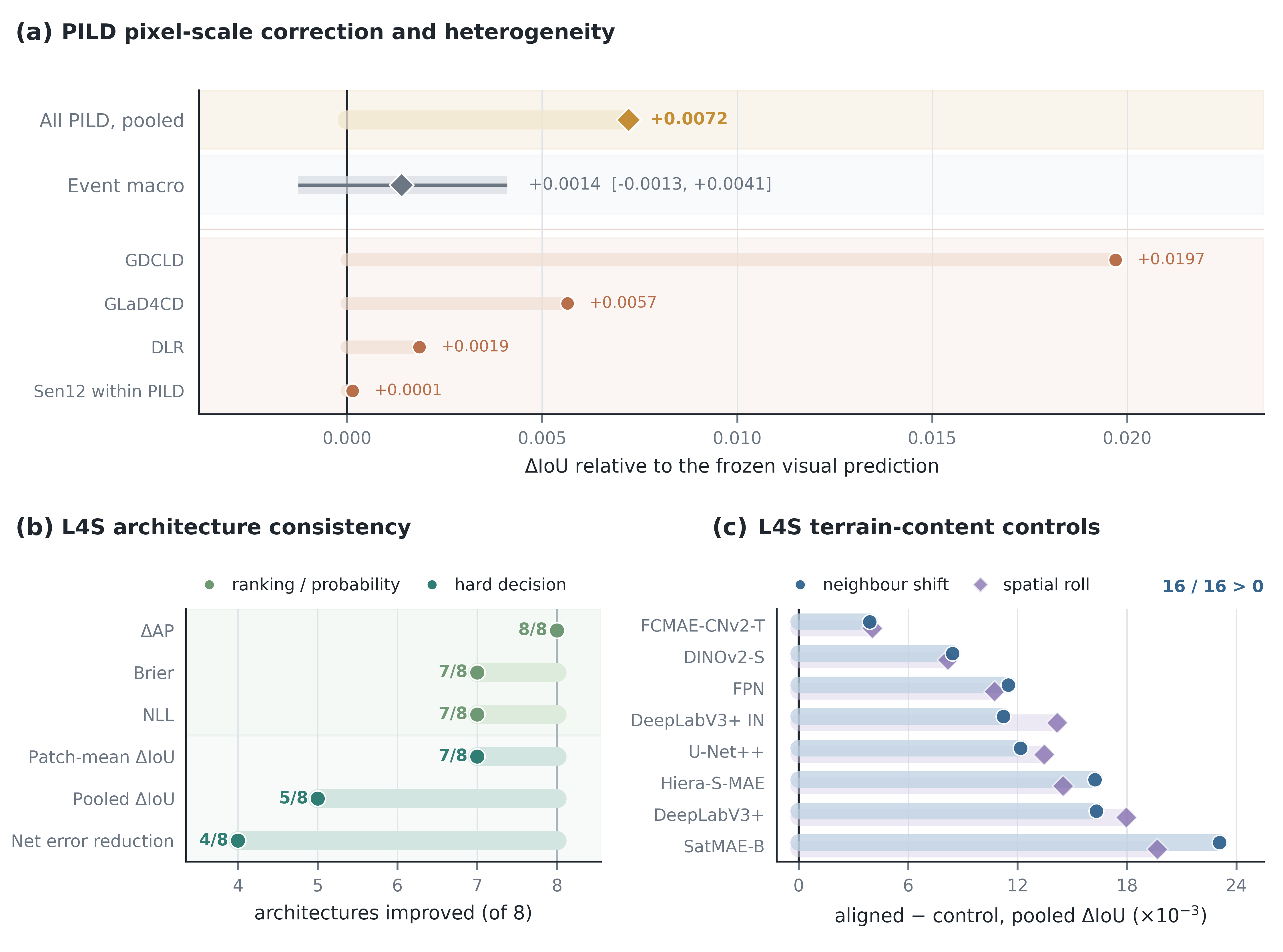}
\caption{Pixel-scale adaptation: net correction, cross-architecture consistency, and terrain-content controls. (a)~$\Delta$IoU for the PILD pool, event macro, and four sources relative to the same frozen visual prediction (Table~\ref{tab:pixel}). (b)~Number of Landslide4Sense architectures improving on six metrics (grey-green: ranking/probability; teal: hard decision). (c)~Pooled foreground IoU increments of aligned terrain versus shift and roll controls across eight architectures (16/16 $>0$). Full per-architecture controls are in Supplementary Table~S4.}
\label{fig:pixel}
\end{figure*}
\dblfloatdefault

\section{Results}

Unless otherwise noted, the main comparisons use the unified PILD corpus of four public sources, 55 canonical events, and 7,890 test samples under four-fold event-isolated cross-prediction, with a frozen Prithvi-EO-2.0-300M-TL visual anchor (pooled baseline IoU 0.21819). The results follow five progressive questions: the object structure of cross-domain visual errors; the corrective reach and scale bottleneck of pixel-level adaptation; native-task information and transfer boundaries of the three priors; object-scale matching with misalignment and optical-competitor attribution; and cross-anchor plus pre-frozen single-shot tests of reproducibility and applicability. Object analysis covers all 55 events that produce candidate bodies, of which 54 contain true-positive pixels for event-macro intervals, together with 62,203 candidate bodies; the pixel- and object-level adaptations share the same corpus, anchor, baseline prediction, and pixel evaluation convention.

\subsection{Cross-Domain Visual Errors Have Object Structure}

\paragraph{Objectified false-positive mass.}
Cross-domain visual errors do not appear only as scattered noise near a true boundary; they also form connected candidates with complete boundaries that can be reviewed separately (Figure~\ref{fig:errorstruct}a). On the unified PILD corpus, the pooled IoU of the frozen visual anchor is 0.21819, and its candidates contain 1,823,755 true-positive and 3,942,700 false-positive pixels, together with a further 2,592,046 missed pixels. The false-positive mass is 2.16 times the true-positive mass and accounts for 60.3\% of all erroneous pixels (FP+FN). The main bottleneck of cross-domain performance is therefore not only imprecise localization of a landslide boundary, but the fact that the visual model fabricates entire landslide bodies that do not exist.

\paragraph{Near-pure spurious bodies and size filters.}
This objectified failure can be quantified directly. The 6,927 samples that contain a prediction produce 62,203 candidate bodies in total, and those with a purity no greater than 0.10, the near-pure spurious bodies, carry 70.3\% of the entire false-positive mass (Figure~\ref{fig:errorstruct}b). If an area of at least 200 pixels is additionally required, this group still carries 37.5\% of the false-positive mass. Area alone, however, cannot constitute a decision rule: the same area threshold also retains 76.9\% of the true-positive mass and 60.9\% of the false-positive mass (Figure~\ref{fig:errorstruct}c). Correct and spurious detections therefore both cluster into objects, and any subsequent method has to review the content compatibility of a candidate body rather than rely on a size filter alone.

\paragraph{Candidate scale and prior support.}
The spatial scale of the candidates explains further why the decision unit of adaptation has to change. After weighting by pixel mass, the median equivalent diameter of the false-positive candidates is 207~m with an interquartile range of 109--407~m, whereas the median for true-positive candidates is 392~m (Figure~\ref{fig:errorstruct}d). On a 10~m analysis grid, 207~m corresponds to a lateral span of about 21 pixels, to about seven grid spacings of the 30~m terrain, and to the same order as the 250~m native support of Material, while the about 5~km support of Trigger still belongs to the event-level background. The object scale can thus simultaneously aggregate several aligned terrain cells, accept the regional susceptibility modulation of Material, and retain Trigger as an event-level condition, without disguising all three priors as 10~m boundary evidence.

Object structure is not the exclusive property of a single data source. The near-pure spurious bodies of GDCLD and Sen12Landslides carry 81.3\% and 71.3\% of their respective false-positive mass, against 39.7\% for DLR and 27.7\% for GLaD4CD. At the same time, Sen12Landslides alone contributes 71.9\% of the pooled false-positive pixels, which shows that the aggregate figures are sensitive to source composition. Although the near-pure false-positive share of DLR is comparatively low, candidate bodies of at least 200 pixels still carry 98.3\% of its true-positive mass. The candidate body is therefore a decision unit that all four sources can construct, but its error purity and its contribution to the pool have to be bounded by per-source results.

These results rewrite the subsequent question from whether large candidates should be deleted into which candidates lack physical support compatible with landslide causation. A pixel-level correction requires a coarse prior to indicate a local boundary precisely, whereas an object-level review need only judge whether an entire candidate body is compatible with the terrain, the material, and the event background. This lowering of the evidential threshold, rather than the size of the candidate bodies themselves, forms the empirical starting point of the scale-matching adaptation experiments that follow.

\subsection{The Corrective Effect and Stability of Pixel-Scale Adaptation}

\paragraph{Pooled corrective effect.}
On the unified PILD corpus, bounded pixel-level adaptation does change the visual judgment, and the change is predominantly corrective. In the four-fold event-isolated cross-prediction, the adaptation removes a net 507,817 erroneous pixels with a corrected-to-harmed ratio of 8.14:1, a pooled error reduction of 7.76\%, and a pooled IoU gain of $+0.00722$ (Figure~\ref{fig:pixel}a). The five-seed Sen12Landslides experiment independently obtains an error reduction of 15.44\%. These results show that aligned terrain support can produce a clear net correction at the pixel scale, but they do not presuppose that the effect appears with equal magnitude in every event and every data source.

\dblfloatbottom
\begin{table*}
\centering
\small
\caption{Main statistics and independent confirmation of pixel-scale adaptation.}
\label{tab:pixel}
\begin{adjustbox}{max width=\textwidth}
\begin{tabular}{llr}
\hline
Cohort & Metric & Result \\
\hline
PILD four-fold OOF (main) & Net erroneous pixels removed; corrected / harmed & \textbf{507,817}; \textbf{8.14 : 1} \\
PILD four-fold OOF (main) & Error reduction; pooled $\Delta$IoU & \textbf{7.76\%}; $\mathbf{+0.00722}$ \\
PILD four-fold OOF (main) & Event-macro $\Delta$IoU [95\% CI] & $+0.00140$ $[-0.00125,+0.00410]$ \\
PILD four-fold OOF (main) & Event composition (positive / zero / negative) & 19 / 26 / 10 \\
Sen12 independent five-seed protocol (confirmation) & Error reduction & \textbf{15.44\%} \\
L4S strict three controls (auxiliary) & Architectures passing the full zero+shift+roll gate (threshold-free / hard decision) & 6 / 8; 7 / 8 \\
\hline
\end{tabular}
\end{adjustbox}
\end{table*}
\dblfloatdefault

\paragraph{Event- and source-level heterogeneity.}
The pooled improvement, however, does not translate proportionally into a cross-event-stable gain in hard decisions. The event-macro $\Delta$IoU is $+0.00140$ with a 95\% confidence interval of $[-0.00125,+0.00410]$ that crosses zero. Of the 55 events, 19 have a positive $\Delta$IoU and 10 a negative one, while a further 26 are exactly zero. A zero is not a missing value but an event in which the adaptation changed no pixel decision, so that its output coincides with the visual prediction; by the operator structure of Section~4.1, the update is identically zero wherever terrain support is invalid. Per-source results show the heterogeneity of the effect further: the pooled $\Delta$IoU of GDCLD, GLaD4CD, DLR, and Sen12Landslides is $+0.0197$, $+0.0057$, $+0.0019$, and $+0.0001$ respectively (Figure~\ref{fig:pixel}a). The net correction over PILD as a whole is therefore real, but its magnitude is markedly conditional on the event and on the source.

\paragraph{Cross-architecture consistency.}
The cross-architecture replay provides complementary evidence. Across the eight vision architectures of the auxiliary Landslide4Sense cohort, average precision improves in every case, and the Brier score and the negative log-likelihood each improve in 7 of 8. When the readout turns to thresholded hard decisions, the sample-mean foreground IoU, the pooled foreground IoU, and the net error reduction improve in 7 of 8, 5 of 8, and 4 of 8 respectively (Figure~\ref{fig:pixel}b). This is not an attempt to force different metrics into a single effect size; it shows instead that the influence of physical support is most consistent on ranking and probability, and becomes more dependent on the specific architecture and sample once it has to translate into a discrete boundary gain. The architectures with a positive net error reduction are divided equally between self-supervised foundation models and modern segmentation configurations, so the phenomenon is not caused by a single architectural family. Table~\ref{tab:pixel} summarizes the main statistics and the independent confirmation.

\noindent Note: The Sen12 row comes from an independent five-seed protocol and its metric is the error reduction, which is not the same as the $\Delta$IoU of the Sen12 source slice of the PILD four-fold OOF in Figure~\ref{fig:pixel}a. Landslide4Sense is an auxiliary cohort used only for cross-architecture consistency and content controls, and does not support a claim of cross-event generalization (see Section~3.4); the six improvement counts across the eight architectures and the 16 aligned-versus-misaligned differences are given in Figure~\ref{fig:pixel}b--c, and the complete architecture-level zero, shift, and roll controls in Supplementary Table~S4.

\paragraph{Aligned-terrain content controls.}
A limited gain in hard decisions does not mean that the physical content is ineffective. Across the eight architectures, the pooled foreground IoU increment of aligned terrain relative to a neighbor-sample shift and to a spatial roll is positive in every case (16/16, Figure~\ref{fig:pixel}c). All controls hold the visual prediction, the adaptation capacity, and the evaluation protocol unchanged and destroy only the spatial correspondence between terrain and sample, so the result attributes the change in prediction to correctly aligned terrain content rather than to additional model capacity. The standardized zero-terrain control and the complete per-architecture values are given in Supplementary Table~S4.

Together, these results reveal a bottleneck of conversion. Physical content can change the probability ranking and produce a net improvement in error flow, but a coarse prior struggles to localize a 10~m segmentation boundary stably. A 250~m Material grid and an approximately 5~km Trigger product are even less able to supply boundary evidence equivalent to a pixel label, whereas Section~6.1 has already shown that the dominant visual errors are organized as complete candidate bodies of about 207~m. The subsequent tests therefore first confirm the information content of the three priors on their native tasks (Section~6.3), and then move the physical decision from the pixel to the candidate landslide body (Section~6.4).

\begin{figure*}[!t]
\centering
\includegraphics[height=0.9\textheight,keepaspectratio]{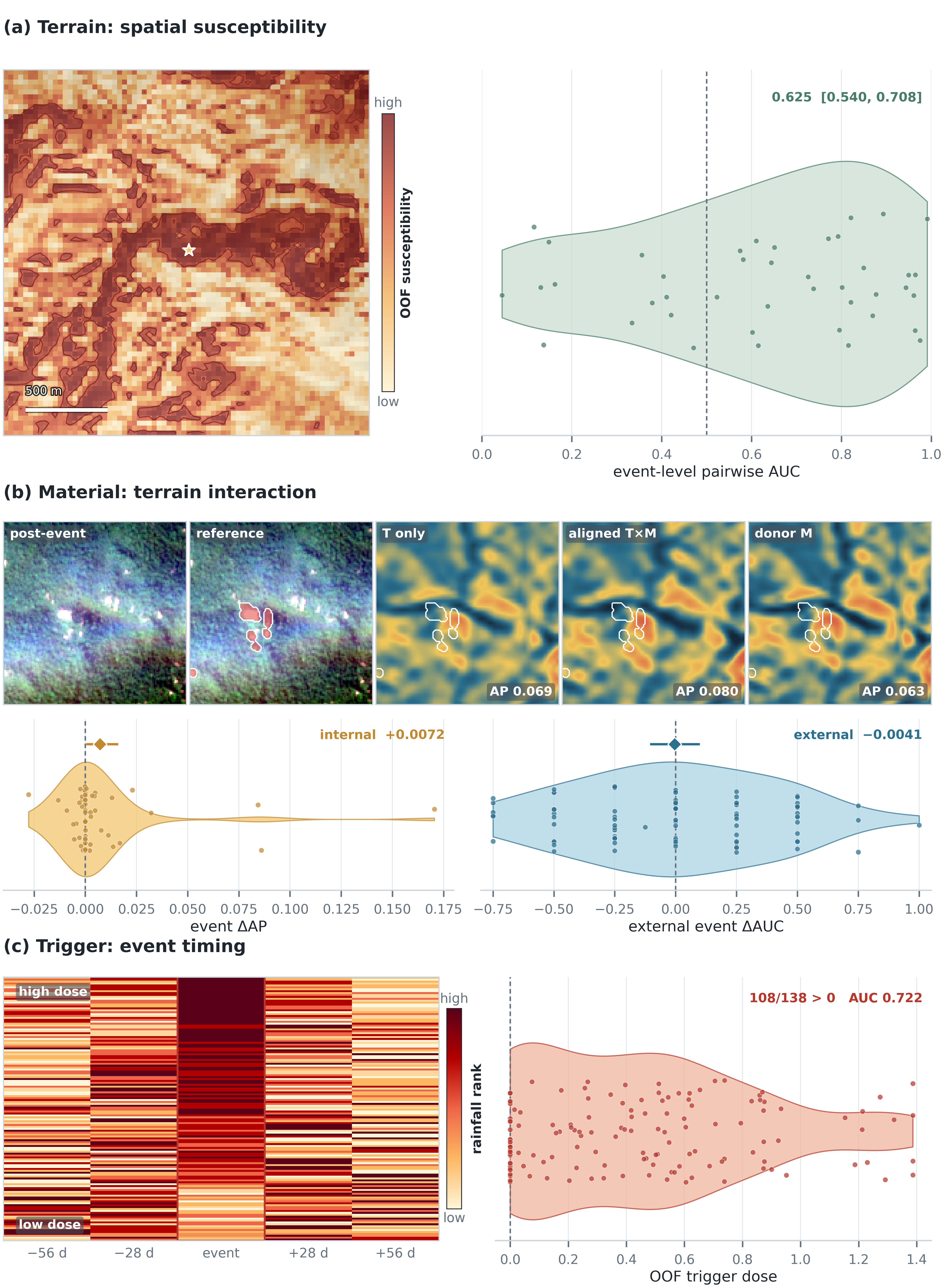}
\caption{Native-task evidence and cross-scale boundary of the three priors. (a)~Terrain: representative out-of-fold susceptibility field and pairwise AUC over 42 spatially isolated events (chance at 0.5). (b)~Material: aligned terrain$\times$material versus a cross-event donor on one sample, with internal AP and external AUC differences. (c)~Trigger: within-event rainfall ranking for the true window versus time-shifted windows, with trigger-dose distribution. Complete statistics and selection rules are in Supplementary Table~S6.}
\label{fig:native}
\end{figure*}

\subsection{Native Information and Transfer Boundary of the Three Physical Priors}

A limited pixel-level gain may arise from two different mechanisms: either the physical variable carries no information, or the information exists but fails to be converted into a stable segmentation intervention across datasets and across scales. To separate the two, Figure~\ref{fig:native} shows directly the real observations, the event-wise distribution, and the independent controls of the three priors on role-matched tasks, while the adjudication of their cross-scale conversion combines the pixel content controls of Section~6.2 with Supplementary Table~S6. None of the three tests uses the visual segmentation output, and each adopts an estimator consistent with its own physical role, so each is compared only against its own reference level and the three are not ranked against one another by effect size.

\paragraph{Native terrain ranking and bounded conversion.}
Terrain forms the most complete chain of evidence. On the GLaD cohort, spatially isolated from the PILD development events, a 100~km distance exclusion and a CopDEM tile exclusion retain 42 events and 38 dependency blocks. The event-balanced pairwise area under the ROC curve (AUC) of the six-feature terrain model is 0.6255, with event bootstrap and dependency-block bootstrap 95\% intervals of $[0.5398,0.7076]$ and $[0.5357,0.7119]$ respectively, both lower bounds above chance. Combined with the consistent advantage of aligned terrain over the shift and roll controls in Section~6.2, terrain has both an independent spatial ranking ability and the capacity to convert into a content-dependent pixel correction; the event-macro interval nevertheless still crosses zero, so its pixel-level conversion is adjudicated as bounded rather than general.

\paragraph{Internal material interaction without external replication.}
Material provides within-dataset interaction evidence but does not pass external transfer. On a susceptibility interaction probe with no optical input and a fixed terrain logit, across 56 source-level events and 7,916 samples (the pre-merge inventory; the primary corpus remains 55 canonical events / 7,890 samples after the single cross-source merge in Supplementary Table~S1), aligned material improves AP by $+0.00719$ (95\% interval $[+0.00102,+0.01549]$) and AUC by $+0.01384$ ($[+0.00054,+0.02929]$) relative to terrain alone; when the same model is given a cross-event material donor at test time only, the aligned input still retains an AP advantage of $+0.00741$ ($[+0.00050,+0.01672]$). On an independent cohort of 92 events and 43 dependency blocks, however, the pairwise AUC difference of adding material to terrain is $-0.0041$ with a block bootstrap interval of $[-0.100,+0.095]$. The design and balance criteria of this test pass 6 of 6, whereas the effect and robustness criteria pass 0 of 10, which shows that the external non-replication cannot simply be attributed to an insufficient sample size or an inadequate test design.

\paragraph{Event-timing trigger evidence without pixel boundaries.}
Triggering provides clear event-timing information but is not boundary evidence at the pixel level. The 138 independent events are divided into 90 storm clusters and cross-predicted at the cluster level; the pairwise AUC of the true pre-event window against four time-shifted windows at the same location is 0.72192, and 108 of 138 events receive a positive trigger dose. In a separate three-window same-season cross-year control (the event year versus the same calendar dates in the previous and following years; chance top-rank rate $1/3$), event-period rainfall ranks first in 75 of 138 events. Although the temporal attribution is stable, the aligned trigger dose does not stably outperform the time-shifted controls when used as a modulation term for pixel segmentation, so the present evidence supports only its event-level dose role and not a direct delineation of 10~m landslide boundaries.

\paragraph{Asymmetric conversion pathways.}
The three priors thereby exhibit asymmetric conversion pathways. Terrain is supported at all three levels of native information, external transfer, and pixel conversion, although the last is only a partial conversion. Material is positive on the internal native task and, after the external cohort failed to replicate, makes no further pixel-level claim. The temporal attribution of Trigger holds, but its external transfer has not been tested independently and the pixel-level Trigger effect likewise abstains. The complete sensitivity analyses, event distributions, external promotion criteria for material, and reproduction entry points are given in Supplementary Table~S6. This result confines the three inputs to distinct roles, with terrain providing dense direction, material providing regional modulation, and triggering providing an event dose, and leads directly to the test of the next section: when coarse physical support cannot localize a single pixel, can raising the decision unit to the candidate landslide body release its corrective value?

\subsection{Object-Scale Matching Amplifies Attributable Physical Correction}

Section~6.2 has shown that aligned terrain can change pixel predictions but that its hard-decision gain is limited by the scale at which it is broadcast, and Section~6.3 has shown that each of the three priors carries information at its own native scale yet cannot be interpreted equivalently as 10~m boundary evidence. The key mechanism tested in this section is therefore not whether adding a physical variable improves the score again, but whether, under the same test corpus, the same frozen visual prediction, and the same evaluation convention, moving the physical decision unit from the pixel to the complete candidate body can substantially widen the net correction.

We define the connected components of the visual prediction as the object-level decision unit, and each candidate body can only be retained or vetoed as a whole; a vetoed region is restored exactly to the visual negative class, and the physical stage adds no new foreground pixel. Of the 7,890 test samples, 6,927 produce at least one predicted candidate body, and a further 963 contain no predicted foreground and pass through unchanged; the former form 62,203 candidate bodies in total. Each candidate body is characterized by a 92-dimensional descriptor composed of terrain geometry, visual confidence, spectral change, and catchment hydrological topology, and the purity prediction uses event-grouped five-fold cross-fitting, so that every candidate body is scored by a model that has not seen the event to which it belongs. The source-conditioned configuration additionally uses only the data-source identifier known at deployment time.

The complete derivation of the criterion is given in the method section. Its core is that, for a candidate body containing $i$ true-positive and $f$ false-positive pixels, vetoing it as a whole can improve the baseline IoU if and only if
\begin{equation}
\begin{array}{@{}l@{}}
\dfrac{i}{f}<\mathrm{IoU}_{\mathrm{base}}
\quad\Longleftrightarrow\quad
\pi<\dfrac{\mathrm{IoU}_{\mathrm{base}}}{1+\mathrm{IoU}_{\mathrm{base}}},
\end{array}
\end{equation}
where $\pi=i/(i+f)$ is the purity of the candidate body. The analytic threshold corresponding to the main visual anchor is 0.17911. This threshold is determined by the baseline performance of the anchor itself and contains no tunable test-set hyperparameter; the pre-frozen re-execution of Section~6.5 further determines this reference quantity from the fitting partition alone. The object-scale main results and the same-corpus pixel contrast are summarized in Table~\ref{tab:object}.

\begin{table*}[!t]
\centering
\small
\caption{Main development-stage results of object-scale physical review, and the same-corpus contrast with the pixel scale.}
\label{tab:object}
\begin{adjustbox}{max width=\textwidth}
\begin{tabular}{lrrr}
\hline
Readout & Source-conditioned & No source identifier & Pixel scale (same corpus, same anchor) \\
\hline
Baseline IoU & 0.21819 & 0.21819 & 0.21820 \\
IoU after review & \textbf{0.24914} & 0.24486 & 0.22542 \\
$\Delta$IoU & $\mathbf{+0.03095}$ & $+0.02667$ & $+0.00722$ \\
Error reduction & \textbf{23.99\%} & 23.36\% & 7.76\% \\
Corrected / harmed ratio & \textbf{9.92} & 8.64 & 8.14 \\
Candidate veto accuracy & 0.891 & 0.882 & --- \\
Share of false-positive mass cleared & 44.2\% & 43.8\% & --- \\
Share of true-positive mass lost & 9.6\% & 10.9\% & --- \\
Purity ranking correlation $\rho$ & 0.463 & 0.444 & --- \\
Event-macro $\Delta$IoU [95\% CI] & $+0.01086$ $[-0.00455,+0.02770]$ & $+0.01196$ $[+0.00020,+0.02632]$ & $+0.00140$ $[-0.00125,+0.00410]$ \\
Event-macro error reduction [95\% CI] & $+12.29\%$ $[+7.08\%,+17.99\%]$ & $+13.27\%$ $[+8.31\%,+18.96\%]$ & $+3.27\%$ $[+1.84\%,+4.91\%]$ \\
Events with positive $\Delta$IoU & 37 / 54 & 32 / 54 & 19 / 55 \\
\hline
\end{tabular}
\end{adjustbox}
\end{table*}

\begin{figure*}[!t]
\centering
\includegraphics[width=\textwidth]{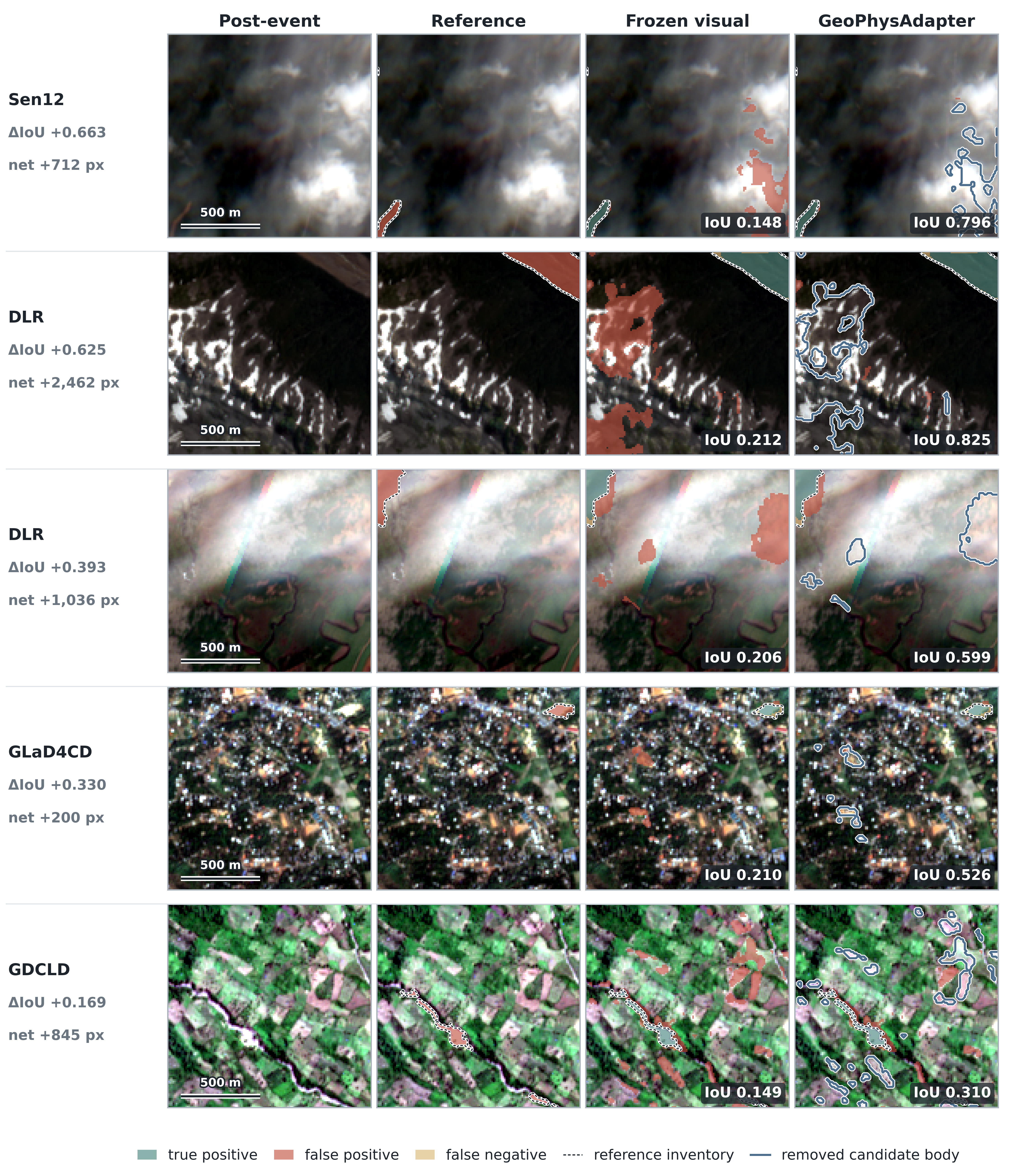}
\caption{High-effect map-level cases of object-scale physical review. Columns: post-event imagery, reference inventory, frozen visual prediction, and GeoPhysAdapter output. Teal/coral/pale gold denote TP/FP/missed pixels; slate outlines mark wholly vetoed candidates. Each panel reports sample-level IoU, $\Delta$IoU, and net corrected pixels; five rows span four sources and five events (scale bar 500~m, full $128\times128$ tiles). Cases illustrate the positive tail only; corpus-level effects are in Table~\ref{tab:object}, and behavioral boundaries in Supplementary Figure~S1.}
\label{fig:objectcases}
\end{figure*}

\noindent Note: The three columns share all 7,890 test samples, the same frozen visual anchor, the same baseline prediction, and the same pixel-level evaluation convention. Object-level adaptation executes a decision only on the 6,927 samples that produce candidate bodies, and the remaining 963 samples revert exactly to the baseline; 62,203 candidate bodies are reviewed in total, of which the source-conditioned configuration vetoes 36,126. All 55 events produce candidate bodies but only 54 contain true-positive pixels, so the object-level event-macro intervals are based on those 54 events, whereas the pixel-level statistics cover all 55. This comparison tests the effect of the operating scale and does not claim that the two adapters share the same internal tensor implementation.

Object-scale matching converts the physical information already identified in the two preceding sections into the strongest corrective effect in the paper. The pooled IoU rises from 0.21819 to 0.24914, an absolute increment of $+0.03095$, and the error reduction, which corresponds more directly to the mechanistic question of this paper, reaches 23.99\%. Relative to pixel-level adaptation on the same corpus, the same anchor, and the same baseline, the error reduction widens from 7.76\% to 23.99\%, a factor of 3.09, and the IoU increment widens from $+0.00722$ to $+0.03095$, a factor of 4.29. The visual prediction, the physical data, and the final pixel evaluation all remain unchanged, and the main difference between the two adaptations lies in the aggregation of physical evidence and in the decision unit; this strictly matched comparison therefore supports scale matching as the principal explanation of the amplification, rather than a change of visual model or an addition of input information.

The error flow shows that this gain is not an indiscriminate contraction of the foreground prediction. The source-conditioned configuration clears 44.2\% of the false-positive mass while losing only 9.6\% of the true-positive mass, with a corrected-to-harmed pixel ratio of 9.92:1 and a candidate veto accuracy of 0.891. Removing the data-source identifier still yields an IoU increment of $+0.02667$ and an error reduction of 23.36\%, which shows that source identity is not the main effect. The event-macro error reduction is $+12.29\%$ with a 95\% interval of $[+7.08\%,+17.99\%]$, whereas the interval of the event-macro $\Delta$IoU crosses zero, indicating that the pixel-mass-weighted overall gain cannot be extrapolated to an equal improvement in every event; this heterogeneity is examined further in Section~6.5.

\paragraph{Attribution to physical content.}
Objectification by itself is not automatically equivalent to a geophysical contribution. To separate generic object post-processing from the net increment of aligned physical content, we use two orthogonal sets of tests, a spatial misalignment intervention and a strong optical competitor, and combine them with the pre-frozen re-execution of Section~6.5 to test whether the attribution is preserved.

\paragraph{Spatial alignment dose response.}
The intervention changes only the terrain stack used to review the candidate bodies, holding the candidates, the spectral descriptors, and the visual confidence unchanged; after the terrain is shifted or replaced as a whole, both the terrain and the hydrological descriptors are recomputed. The corrective effect at the development stage decays monotonically with the degree of misalignment: the $\Delta$IoU of aligned terrain is $+0.03095$, falling to $+0.01539$ after a 320~m shift, to $+0.01358$ after a 640~m shift, and to only $+0.00798$ for a same-source cross-event donor, while the purity ranking correlation falls in step from 0.463 to 0.350, 0.341, and 0.289. Aligned terrain retains $+0.01556$ relative to the closest misaligned condition, the 320~m shift, and $+0.02297$ relative to the cross-event donor.

This ordering is preserved on the pre-frozen held-out partition, where the $\Delta$IoU of the aligned, 320~m shifted, 640~m shifted, and cross-event donor conditions is $+0.02667$, $+0.02022$, $+0.01921$, and $+0.01408$ in turn, and aligned terrain retains $+0.00645$ relative to the 320~m shift and $+0.01259$ relative to the cross-event donor. That the misalignment controls remain positive is not a counterexample but a revelation that the total gain contains two parts: object-level decision making and visual confidence can supply a generic reviewing ability, and only the stable margin of the aligned condition over the misaligned conditions constitutes a location-specific geophysical contribution.

\paragraph{Contrast with a strong optical competitor.}
An alternative explanation is that the same gain could be obtained by a sufficiently strong appearance reviewer. We therefore construct a 39-dimensional object-level spectral and change descriptor covering the Sentinel-2 bands, common spectral indices, pre-post differences, within-body heterogeneity, and candidate-to-ring contrast. All arms share the candidate bodies, the learner, the seeds, the cross-fitting partition, and the analytic criterion, and the only thing that varies is the descriptor family; the full six-arm ablation is reported in Supplementary Table~S5d.

The complete spectral-change evidence plus visual confidence yields an IoU increment of only $+0.0037$, whereas terrain, spectra, and confidence together reach $+0.0232$; relative to the strong optical competitor, aligned geophysical content therefore provides an independent increment of $+0.01945$. Replacing the aligned spectra with within-source cross-event spectra lowers the gain from $+0.0232$ to $+0.0199$, which shows that spectra offer only a minor supplement. Although adding spectra raises the overall ranking correlation from 0.376 to 0.427, the correlation for large candidates rises only from 0.476 to 0.487 and does not further increase the deployable IoU. The final gain therefore depends on ranking correctly those large candidates that carry the dominant error mass, rather than on raising the average correlation over all candidates.

An independent catchment-hydrology ablation shows further that the physical contribution is not confined to local slope form. The 23-dimensional catchment hydrological descriptor plus visual confidence obtains a deployable $\Delta$IoU of $+0.01648$ with a ranking correlation of 0.379, whereas the 27-dimensional terrain geometry stack plus visual confidence gives $+0.02344$ and 0.373. Terrain geometry produces the larger final correction, yet the hydrological topology, which contains no local form of the candidate body, still retains an independent capability, which shows that slope geometry and the position of a candidate within the drainage system constitute complementary evidence.

The three sets of results thus close the attribution. The configuration without a source identifier excludes the explanation by data-source identity, the aligned-misaligned ladder identifies the location-specific physical increment, and the strong optical competitor excludes the explanation by appearance post-processing alone. Not all of the object-scale gain comes from physics, but one part of it is reproducible, interventionable, and preserved across the held-out partition, and that part depends explicitly on aligned terrain and hydrological content.

\paragraph{High-effect mapping cases and behavioral boundaries.}
Figure~\ref{fig:objectcases} serves only to display the map-level form of correction that the mechanism can achieve in the positive tail, and not to estimate the overall effect or its frequency of occurrence. Following a recomputable rule, we take from each of the four data sources the sample with the highest $\Delta$IoU that satisfies the area and actual-veto conditions, and then add the highest-effect sample from one further distinct event; the five cases come from five events, with sample-level $\Delta$IoU of $+0.663$, $+0.625$, $+0.393$, $+0.330$, and $+0.169$. They show that GeoPhysAdapter can clear as a whole the spatially coherent candidates produced by the frozen visual model while retaining the reference landslide areas. The overall effect remains that of the full-corpus statistics and event resampling in Table~\ref{tab:object}: of the 6,927 samples containing a prediction, 5,264 show a net reduction in error, 1,198 are unchanged, and 465 show a net harm, and at the event level 44 of 55 are net positive. Representative boundaries of collateral harm, net harm, and an unchanged identity fallback are given in Supplementary Figure~S1.

\subsection{Mechanism Reproduction and an Operationalized Test of the UGCoP}

\begin{figure*}[!t]
\centering
\includegraphics[width=\textwidth]{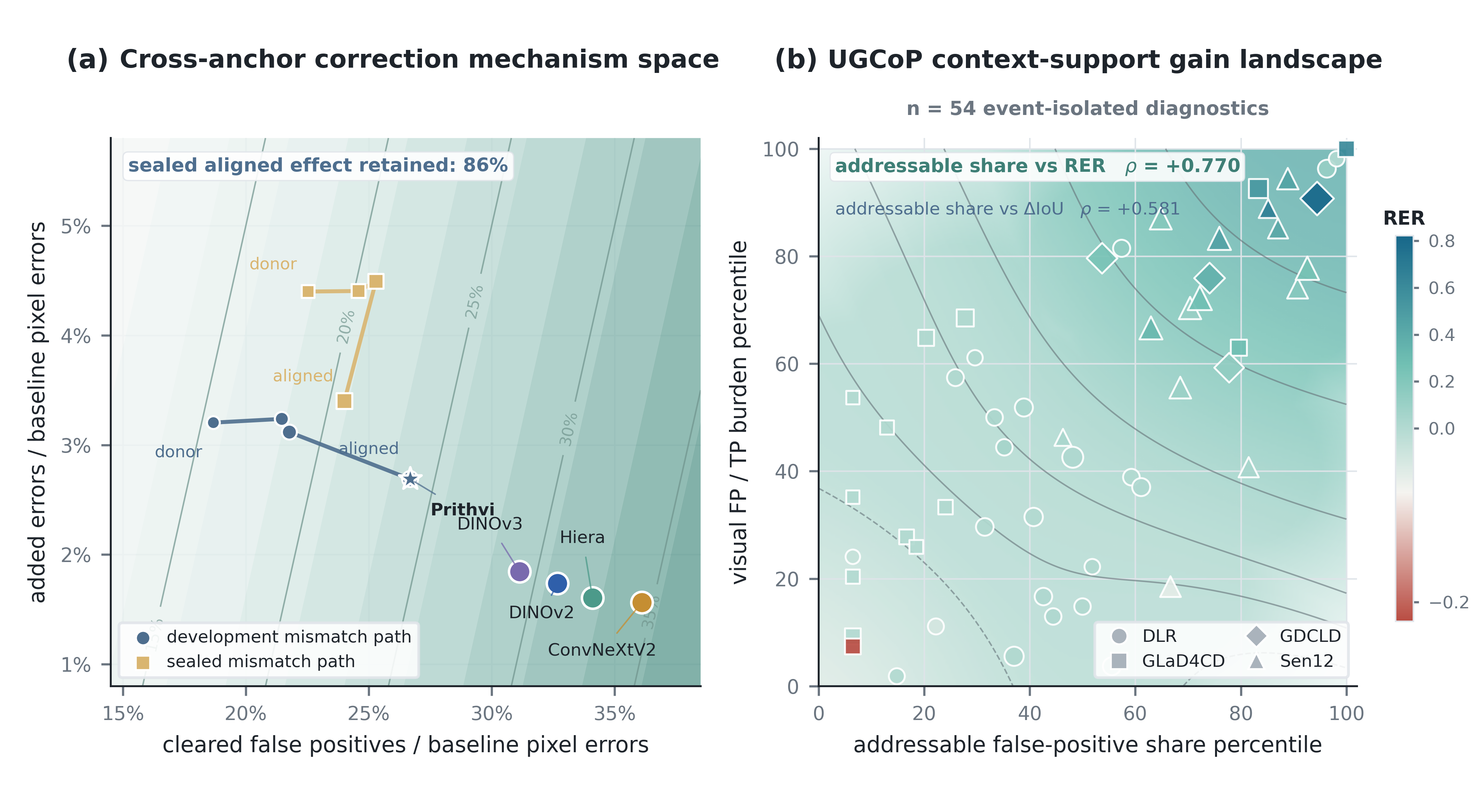}
\caption{Cross-anchor mechanism reproduction and the UGCoP context-support landscape. (a)~Correction--harm space of five frozen visual anchors; trajectories move from aligned terrain to shift, roll, and cross-event donor (held-out aligned effect retains 86\% of development). (b)~Event-level landscape of addressable false-positive share versus FP/TP burden for 54 events, with marker color denoting error reduction. Contours are descriptive; inference uses the overlaid raw events (Table~\ref{tab:anchors}; Supplementary Tables~S5a--S5c).}
\label{fig:ugcop}
\end{figure*}

If object-scale matching reflects a stable mechanism rather than an accident of one visual model or one development process, it should satisfy three conditions at once: it should preserve the direction of error reduction on different visual anchors, it should retain the main effect after the analytical degrees of freedom have been frozen, and the between-event gain should be explicable by a predefined structure of visual error. We accordingly organize a cross-anchor reproduction, a pre-frozen single-shot re-execution, and an event-level mechanistic diagnosis; Figure~\ref{fig:ugcop} unifies the three into two complementary views, a correction-harm mechanism space and an event-level gain landscape.

\paragraph{Cross-anchor reproduction.}
We evaluate Prithvi-EO-2.0, DINOv3-SAT-L, Hiera-S-MAE, ConvNeXtV2-FCMAE, and DINOv2-S under the same data contract. The five anchors share the test samples, the event-isolated split, the sampling scheme, the optimization budget, and the validation threshold rule, and the only thing that changes is the frozen visual encoder. Each anchor uses its own pooled baseline IoU in the analytic criterion, so this experiment tests whether the direction of the effect and the error flow are preserved, and does not average increments that lack a common reference. Results for the five anchors are reported in Table~\ref{tab:anchors}.

\begin{table*}[!t]
\centering
\small
\caption{Object-level review on five visual anchors, each with its own baseline.}
\label{tab:anchors}
\begin{adjustbox}{max width=\textwidth}
\begin{tabular}{llrrrr}
\hline
Visual anchor & Pretraining source & Own baseline IoU & $\Delta$IoU (source-conditioned) & Error reduction & $\Delta$IoU (no source identifier) \\
\hline
Prithvi-EO-2.0-300M-TL & Remote sensing & 0.21819 & $\mathbf{+0.03095}$ & \textbf{23.99\%} & $+0.02667$ \\
DINOv3-SAT-L & Satellite imagery, self-supervised & 0.11290 & $\mathbf{+0.02048}$ & 29.29\% & $+0.01871$ \\
Hiera-S-MAE & General natural imagery & 0.10093 & $\mathbf{+0.02382}$ & 32.50\% & $+0.02328$ \\
ConvNeXtV2-tiny-FCMAE & General natural imagery & 0.09076 & $\mathbf{+0.02319}$ & 34.55\% & $+0.02308$ \\
DINOv2-S & General natural imagery & 0.09497 & $\mathbf{+0.01756}$ & 30.94\% & $+0.01849$ \\
\hline
\end{tabular}
\end{adjustbox}
\end{table*}

\noindent Note: The baselines of the four alternative anchors are lower than that of the main Prithvi anchor, so they have a larger clearable false-positive space. Each anchor uses its own baseline, and gains are not averaged across anchors.

All five anchors produce a positive IoU increment, ranging from $+0.01756$ to $+0.03095$, with an error reduction of 23.99\% to 34.55\%. More importantly, Figure~\ref{fig:ugcop}a projects the five anchors into a unified correction-harm mechanism space, in which the horizontal axis is the share of cleared false positives in the pixel error before adaptation, the vertical axis the share of newly introduced error, and the analytic contours the difference between the two. All anchors lie in the region where the amount cleared is substantially greater than the amount harmed, with a corrected-to-harmed ratio of 9.92 to 23.08. The satellite-imagery self-supervised DINOv3-SAT-L obtains an IoU increment of $+0.02048$ and an error reduction of 29.29\%, which shows that the mechanism depends neither on natural-image pretraining nor on a single Prithvi representation; the main effect size remains anchored on Prithvi, which has the strongest baseline.

\paragraph{Pre-frozen single-shot re-execution.}
The development results use event-grouped cross-prediction, but the model configuration has itself undergone selection. To constrain the freedom of repeated runs and configuration adjustment, we freeze the descriptors, the learner, the five seeds, the analytic criterion, the partition rule, and the input hashes before execution, and then divide the 55 events into 26 fitting events and 29 held-out events by an external deterministic rule. The model and the threshold are determined by the fitting partition alone, and the held-out statistics take no part in selection.

The $\Delta$IoU of the held-out partition is $+0.02667$ and the error reduction 20.61\%, both retaining about 86\% of the main development effect. The development and held-out trajectories in Figure~\ref{fig:ugcop}a both move from the aligned condition toward the 320~m shift, the 640~m roll, and the cross-event donor, and the net corrective space contracts as misalignment increases; under the held-out condition, the aligned case still retains $+0.00645$ relative to the closest misaligned condition and $+0.01259$ relative to the cross-event donor. This execution serves to constrain the analytical degrees of freedom rather than to claim prospective independent validation, because the labels of these events were accessed during earlier data development. The complete values, the provenance of the threshold, and the whole-source holdout results are given in Supplementary Table~S5a.

\paragraph{Gain concentration and the boundary of applicability.}
The object-level gain is highly heterogeneous across events. If its main action is to clear near-pure spurious bodies, then the addressable false-positive mass predefined within an event should predict the gain. Figure~\ref{fig:ugcop}b uses the within-event percentile of the addressable false-positive share and of the false-positive to true-positive burden as two coordinates, and displays the actual error reduction of the 54 events. To render the continuous trend, the background uses a Gaussian kernel regression whose bandwidth is selected by leave-one-event-out error, with transparency controlled by the effective support; all raw events remain overlaid, and statistical inference uses the raw events alone and not the smoothed surface. The Spearman correlation between the addressable share and the error reduction is $\rho=+0.770$ ($p=9.9\times10^{-12}$), and that with $\Delta$IoU is $+0.581$ ($p=4.1\times10^{-6}$), whereas the correlation using the share of large candidates alone is not significant. The gain therefore depends on how much near-pure spurious mass the visual model has produced that physical evidence can identify, and not on the size of the candidate bodies themselves. The complete correlation matrix and the per-source decomposition are given in Supplementary Tables~S5b--S5c.

This mechanism also bounds the scope of the claim. The gain concentrates in the events and sources with more addressable error, but the error structure cannot fully explain differences of sensor, annotation, and region. On the pre-frozen held-out partition, the event-macro $\Delta$IoU is $-0.00317$ with a 95\% interval of $[-0.01892,+0.00853]$, while the event-macro error reduction remains $+8.55\%$ with an interval of $[+2.37\%,+15.86\%]$; a significant decrease in pooled error therefore does not mean that every event improves in IoU. Under a whole-source holdout, the purity correlation falls to 0.242 and the $\Delta$IoU is only $+0.0006$, so the present evidence supports only transfer to new events within a known source and not a zero-shot claim for an entirely new data source. The segmentation abstention boundaries of Material and Trigger have already been given in Section~6.3 and are not repeated here.

\paragraph{An empirical test of the UGCoP.}
In this paper, the UGCoP is operationalized as the uncertain correspondence between the geographic context available to a model and the true hazard-generating context that produced the current visual error. Three results jointly support this interpretation. Aligned terrain has a higher corrective effect than the shift, the roll, and the cross-event donor, which shows that contextual content in the correct location cannot be replaced by an arbitrary background. The event gain increases with the addressable physical support, which shows that context converts into correction only where it intersects the structure of visual error. And the effect weakens markedly under a whole-source holdout, which reveals that this correspondence remains constrained by sensor, annotation system, and regional background. We therefore do not treat the UGCoP as an abstract backdrop, but extend it into a testable proposition: the physical gain in cross-domain landslide mapping depends on whether the content, the location, and the decision scale of the context are matched at the same time. These results provide empirical support for the UGCoP holding within the task of cross-domain inference by a vision foundation model, and at the same time delimit its scope of applicability, rather than stating it as a universal proof for all geographic settings.

Taking the five sets of results together, the scale-matching claim forms a closed chain of evidence. Cross-domain visual errors cluster into near-pure spurious bodies; pixel-level physical adaptation can change the ranking and the error flow, but its hard-decision effect is limited by scale; an object-level decision widens the error reduction to 23.99\% and preserves it across five visual anchors and a pre-frozen single-shot execution; and its gain decays with physical misalignment and increases with the addressable visual error within an event. The adaptivity of GeoPhysAdapter is thus expressed not only in correcting where the evidence is sufficient, but also in retaining the visual prediction where support is inadequate or the role is mismatched. This behavior of intervening when matched and abstaining when mismatched turns the UGCoP from a theoretical backdrop into a testable mechanism in cross-domain visual inference.

\section{Discussion}

This study does not set out to prove that adding a physical channel necessarily raises average IoU. It addresses a more mechanistic cross-domain question instead: when a vision foundation model produces structured errors, can geophysical priors supply an attributable correction at a decision unit commensurate with their own scale of information? The results decompose the answer into three connected judgments. That physical information is genuinely used does not mean that it necessarily brings a beneficial hard decision. That physical information is scientifically relevant does not mean that it can localize a 10~m boundary. And the gain that truly transfers depends on whether the physical support, the visual error, and the decision scale are matched.

\subsection{Scale Matching and Role Constraints Jointly Determine How Physical Information Intervenes}

The same-corpus contrast between the pixel and object levels provides the most direct mechanistic evidence in this paper. The two adaptations share the same test samples, the same visual prediction, and the same pixel evaluation metric, yet object-level adaptation substantially widens the pooled corrective effect (Table~\ref{tab:object}). This difference cannot simply be explained by the object model having more parameters, because aligned terrain shows a monotonic dose response under spatial shift and cross-event donor misalignment, and because the geophysical descriptors still retain an independent increment over a strong optical reviewer (Section~6.4).

The phenomenon originates in the scale relation between visual error and physical support. The dominant problem produced by a cross-domain visual anchor is not isolated pixel noise but near-pure spurious candidate bodies. For errors of this kind, terrain and elevation-derived hydrological information need not recover a fine boundary; they need only judge whether an entire candidate body lies in a hillslope and catchment position compatible with gravity-driven failure. Pixel-level adaptation, by contrast, requires coarse support to give a precise direction for every 10~m pixel, and therefore tends to manifest as an improvement in ranking and in probability mass rather than as a stable and large gain in hard decisions. What this paper calls scale matching is therefore not a generic fusion of multi-scale features, but the correspondence among the support scale of a physical prior, the spatial structure of the visual error, and the final decision unit---the same principle that object-based remote-sensing analysis has long used to argue that the analysis unit determines which contextual evidence can act \citep{Blaschke2010,Martha2010,Ma2017OBIA}. The same physical content yields only a limited gain at the pixel scale yet can reduce error substantially at the scale of the candidate body, which shows that a physical prior is not better for being finer, stronger, or more deeply coupled.

At the same time, terrain, material, and triggering jointly describe the setting in which a landslide occurs, but they do not carry the same segmentation duty---a role separation consistent with susceptibility science and with physics-informed learning more broadly \citep{Reichenbach2018,Reichstein2019,Karniadakis2021,Dahal2025}. Terrain has continuous spatial structure and is the only prior permitted to produce a dense direction of correction; its native-task evidence and the cross-architecture content controls support transferable, location-dependent spatial information (Sections~6.2--6.3). Material, with a support of about 250~m, is better suited to modulating terrain susceptibility than to delineating a boundary independently: the fixed-terrain interaction within PILD supports that modulation, but the effect did not replicate on an independent external cohort and did not convert stably into a segmentation gain. Trigger answers the temporal forcing of an event and can carry temporal discriminability, yet it usually lacks within-sample spatial variation and therefore cannot provide a pixel-level direction. Broadcasting the latter two classes of background onto a 10~m grid adds no physical resolution and only increases the opportunity for source shortcuts.

Active abstention is therefore not a remedial explanation for two of the three priors having failed to contribute; it is part of the methodological contract of GeoPhysAdapter. Terrain supplies a direction where its spatial support is valid, material modulates an existing terrain response only where medium evidence is available, and triggering adjusts the intervention budget only where event-timing evidence is valid; any input whose support is invalid, or which fails a misalignment control, reverts exactly to the identity state. This design keeps the three priors within one scientific framework while avoiding the packaging of information at different scales into three symmetric pixel-level experts (Supplementary Tables~S2a--S2b).

\subsection{Attributable Gain and Executable UGCoP Principles}

That the decision scale matters does not by itself show that the gain comes from physical content. We therefore complete the attribution with three complementary sets of evidence, following the broader lesson that cross-domain gains must be stress-tested against shortcuts and misaligned context rather than accepted from channel addition alone \citep{Geirhos2020,Hong2023,Rafi2024}: (i)~content intervention, in which the visual state and the decision rule are held fixed while only the spatial correspondence of terrain is destroyed, after which the corrective effect decays with misalignment and aligned terrain passes the shift and roll gates across architectures; (ii)~a strong optical competitor, which shows that objectification and spectral or confidence features cannot adequately explain the geophysical increment; and (iii)~a source control, which retains the main effect after the data-source identifier is removed. Cross-anchor reproduction and the pre-frozen single-shot re-execution further show that the mechanism does not depend on a single foundation model or on repeated development-time tuning (Section~6.5; Supplementary Tables~S3--S5).

These tests still do not license an interpretation of the object model as a full simulation of the landslide process. The purity regression estimates how likely a visual candidate is to be real, and the physical descriptors provide evidence of consistency with the conditions of landslide formation rather than a direct inversion of displacement, pore water pressure, or factor of safety. The causal statement this paper can support is that, with the visual state and the decision rule held fixed, destroying the spatial correspondence of physical content systematically weakens the corrective effect. The stronger statement it cannot support is that some descriptor has recovered the true process of failure.

On this basis, the UGCoP notes that the geographic context available to a model need not be the geographic setting that actually acts on the object of study \citep{Kwan2012a,Kwan2012b}. Our results turn this theoretical problem into three executable conditions for trustworthy GeoAI adaptation \citep{Wang2024,Janowicz2020,Hochmair2025}. First, a physical variable must have an auditable provenance and quality state, and resampling must not be misread as an increase in resolution. Second, physical content must be used at a decision unit commensurate with its scale of action. Third, the adapter must be permitted to leave the visual prediction unchanged where support is missing, misaligned, or without increment. Trustworthiness therefore comes not only from a higher pooled score, but from the auditability of what was changed, on what basis, and when nothing was changed. The adapter clears a large share of false-positive mass while retaining an identifiable true-positive cost structure, and its event-level gain correlates with addressable false-positive mass rather than with candidate size itself (Sections~6.4--6.5). This shows that it does not contract predictions universally, but acts where the visual model has produced structured errors that physical support can identify. The conclusion is narrower than the claim that physics always raises average accuracy, yet it has more explicit testable conditions and clearer implications for deployment.

\subsection{Scope and Limitations}

Four boundaries remain important. First, the main object results come from a development-stage out-of-fold evaluation with event grouping. The pre-frozen single-shot re-execution constrains configuration selection and repeated runs, but the labels of the held-out events were accessed during earlier data development, so it cannot be called prospective independent validation. Second, the main source-conditioned configuration targets new events within a known source; under a whole-source holdout the effect is weak and does not yet support a claim of zero-shot deployment on an entirely new data source. Third, the pooled improvement concentrates in events with a large addressable false-positive mass. The event-macro error reduction is stably positive, but the confidence interval of the event-macro $\Delta$IoU can cross zero, and GeoPhysAdapter does not guarantee an IoU improvement in every event. Fourth, the present object-level operation only vetoes visual candidates, which suppresses false positives effectively but cannot generate a new landslide boundary where the visual model has missed one entirely.

Residual-error structure, support coverage, and per-source differences are reported in Supplementary Section~S5 and Supplementary Figure~S1. The data themselves also limit the physical interpretation that can be made: CopDEM terrain at about 30~m is sufficient to support hillslope and hydrological review at the scale of a candidate body, but cannot represent sub-hillslope microtopography; Material remains constrained by external non-replication; and Trigger supports only event-level temporal attribution. Future work needs to carry out a prospective frozen validation on genuinely unvisited new events, to introduce high-resolution terrain and geotechnical variables with independent spatial support, and to extend the present veto of false positives into the physically constrained generation of candidates missed by the visual model. The reproduction information necessary for the method, the data, and the state of the evidence is collected in Supplementary Tables~S1--S6.

\section{Conclusion}

We proposed GeoPhysAdapter in order to answer whether coarse geophysical priors can provide a trustworthy correction when a vision foundation model fails across domains. The core answer is affirmative but conditional: a physical prior converts into a stable segmentation gain only where its provenance is valid, where its spatial or temporal support is credible, and where its scale of action matches the decision unit of the visual error. Terrain can provide a spatial direction, and material and triggering provide the medium and the event background respectively; none of the three can be broadcast indiscriminately as 10~m boundary evidence.

On an event-isolated dataset of four sources, 55 events, and 7,890 samples, pixel-level adaptation removes a net 507,817 erroneous pixels with an error reduction of 7.76\%, and object-scale matching further raises IoU from 0.21819 to 0.24914 and the error reduction to 23.99\%, with a corrected-to-harmed pixel ratio of 9.92:1. The gain decays monotonically with terrain misalignment, still retains an independent IoU increment of $+0.01945$ relative to a strong optical reviewer, and remains positive across five visual anchors and a pre-frozen single-shot re-execution. At the same time, the external non-replication of material, the failure of triggering to convert into a segmentation increment, and the limited gain under a whole-source holdout together delimit the boundary of applicability of the present framework.

These results turn the UGCoP from an interpretation of background uncertainty into an executable methodological principle: the value of physical information is determined not by the number of channels, but jointly by context fidelity, role constraint, scale matching, and active abstention. GeoPhysAdapter therefore does not replace a vision foundation model with a physical model; it applies a limited correction with auditable geophysical evidence where the visual model has produced addressable structured errors. This provides a trustworthy GeoAI pathway for cross-domain landslide mapping that balances performance, attribution, and transparency of boundaries.

\section*{Funding}

This research was supported by a Smart Traffic Fund (PSRI/44/2208/PR) from the Hong Kong Productivity Council and Hong Kong Transport Department, a grant from the 1+1+1 CUHK-CUHK(SZ)-GDSTC Joint Collaboration Fund (4760974, 2025A0505000062), and the Start-Up Grant (SUG) project ``Geospatial Artificial Intelligence for Climate Resilient Urban Environment'' from the National University of Singapore (E-109-00-0036-01).

\section*{Data availability}

The study builds on public source datasets whose raw imagery, labels, and derived physical layers remain under the licenses of their original providers. We therefore release curated metadata, protocol assets, split definitions, source-license notes, download pointers, and release-safe derived files via Zenodo at \url{https://doi.org/10.5281/zenodo.19430714}, rather than redistributing licensed raw imagery.

\section*{Code availability}

Training, evaluation, caching, and figure-generation workflows are available at \url{https://github.com/Liu-Zhihang/geophysadapter}, including the curated scripts, environment specification, and utilities needed to reproduce the reported tables and figures. Files not required for reproduction, such as exploratory utilities and large local caches, are excluded.

\section*{Declaration of competing interest}

The authors declare that they have no known competing financial interests or personal relationships that could have appeared to influence the work reported in this paper.

\bibliographystyle{elsarticle-harv}
\bibliography{reference}

\end{document}